\documentclass{ieeeaccess}
\usepackage{cite}
\usepackage{amsmath,amssymb,amsfonts}
\usepackage{algorithmic}
\usepackage{graphicx}
\usepackage{textcomp}
\usepackage{booktabs}
\usepackage{multirow}
\usepackage{commath}
\usepackage{color}
\usepackage{xspace}

\usepackage{subcaption}
\usepackage{cleveref}
\usepackage{my_commands}
\usepackage{algorithm}
\usepackage{pifont}
\newcommand{\xmark}{\ding{55}}%
\usepackage[all]{nowidow}

\def\BibTeX{{\rm B\kern-.05em{\sc i\kern-.025em b}\kern-.08em
    T\kern-.1667em\lower.7ex\hbox{E}\kern-.125emX}}

\begin{document}

\title{Particle-Filtering-Based Latent Diffusion for Inverse Problems}
\author{\uppercase{Amir Nazemi}\authorrefmark{1},
\uppercase{M. Hadi Sepanj}\authorrefmark{1}\authorrefmark{2}, \uppercase{Nicholas Pellegrino} \authorrefmark{1}\authorrefmark{2},  \uppercase{Chris Czarnecki} \authorrefmark{1}\authorrefmark{2}, and \uppercase{Paul Fieguth} \authorrefmark{1}}

\address[1]{Vision \& Image Processing Lab, Department of Systems Design Engineering, University of Waterloo,\\ Waterloo, Ontario, Canada~(e-mail: amir.nazemi, mhsepanj, nicholas.pellegrino, cczarnecki, paul.fieguth@uwaterloo.ca)}

\address[2]{Equal contribution}
\tfootnote{We thank NSERC Alliance and the Digital Research Alliance of Canada (alliancecan.ca) for their generous support of this research.}

\markboth
{Nazemi \headeretal: Particle-Filtering-Based Latent Diffusion for Inverse Problems}
{Nazemi \headeretal: Particle-Filtering-Based Latent Diffusion for Inverse Problems}

\corresp{Corresponding author: Amir Nazemi (e-mail: amir.nazemi@uwaterloo.ca).}

\begin{abstract}
Latent-diffusion solvers for image inverse problems commonly generate a reconstruction from a single random initialization, making their output sensitive to the sampled trajectory. We introduce particle-filtering-based latent diffusion (PFLD), a framework that evolves multiple latent trajectories during the early stages of reverse diffusion, assigns them Cauchy-inspired measurement-consistency weights, and progressively resamples and prunes the particle population. The particle-management layer can be combined with diffusion-based solvers for linear or nonlinear inverse problems. Using PSLD as the base solver, we evaluate PFLD on eight-fold super-resolution, Gaussian deblurring, and inpainting with the FFHQ-1K and ImageNet-1K datasets at $512\times512$ resolution. Relative to the single-particle PSLD implementation, PFLD improves LPIPS, PSNR, and SSIM for super-resolution and Gaussian deblurring on both datasets and for inpainting on FFHQ-1K; ImageNet-1K inpainting results are mixed. PFLD also improves FID in four of the six task--dataset configurations while particle pruning substantially reduces the cost relative to retaining every trajectory for the full reverse process.
\end{abstract}


\titlepgskip=-21pt

\maketitle

\section{Introduction}
\label{sec1}
In recent years, \textit{generative} methods~\cite{harshvardhan2020comprehensive, compton2013generative} have come into use for solving inverse problems~\cite{rout2023solving, chung2022diffusion,dimakis2022deep}. 
These methods make use of a prior encapsulated in a \textit{generative model} to best solve a given class of inverse problems~\cite{shah2018solving,bertero2021introduction}. 
In particular, the approach of using generative models offers a substantial improvement compared to conventional deterministic methods~\cite{lunz2018adversarial} due to their inherent ability to \emph{sample} from the \emph{posterior-distribution}, thereby yielding more plausible estimates~\cite{fieguth2010statistical}.

From the family of generative models, diffusion models~\cite{sohl2015deep, croitoru2023diffusion,song2020score}, further introduced under Preliminaries, have emerged as a compelling framework. 
Indeed, models such as DPS~\cite{chung2022diffusion}, PSLD~\cite{rout2023solving}, Soft Diffusion~\cite{daras2023soft}, Cold Diffusion~\cite{bansal2024cold}, STSL~\cite{rout2023beyond}, P2L~\cite{chung2023prompt}, and DOC~\cite{li2024solving} have demonstrated impressive performance.
While these methods show promise, output quality can vary with random initialization, as illustrated for PSLD in \Cref{fig:psld_10_results}. Several single-trajectory methods do not explicitly compare alternative initial trajectories before committing computational effort to a reconstruction.


\begin{figure}[t]
    \centering
    \includegraphics[width=\linewidth]{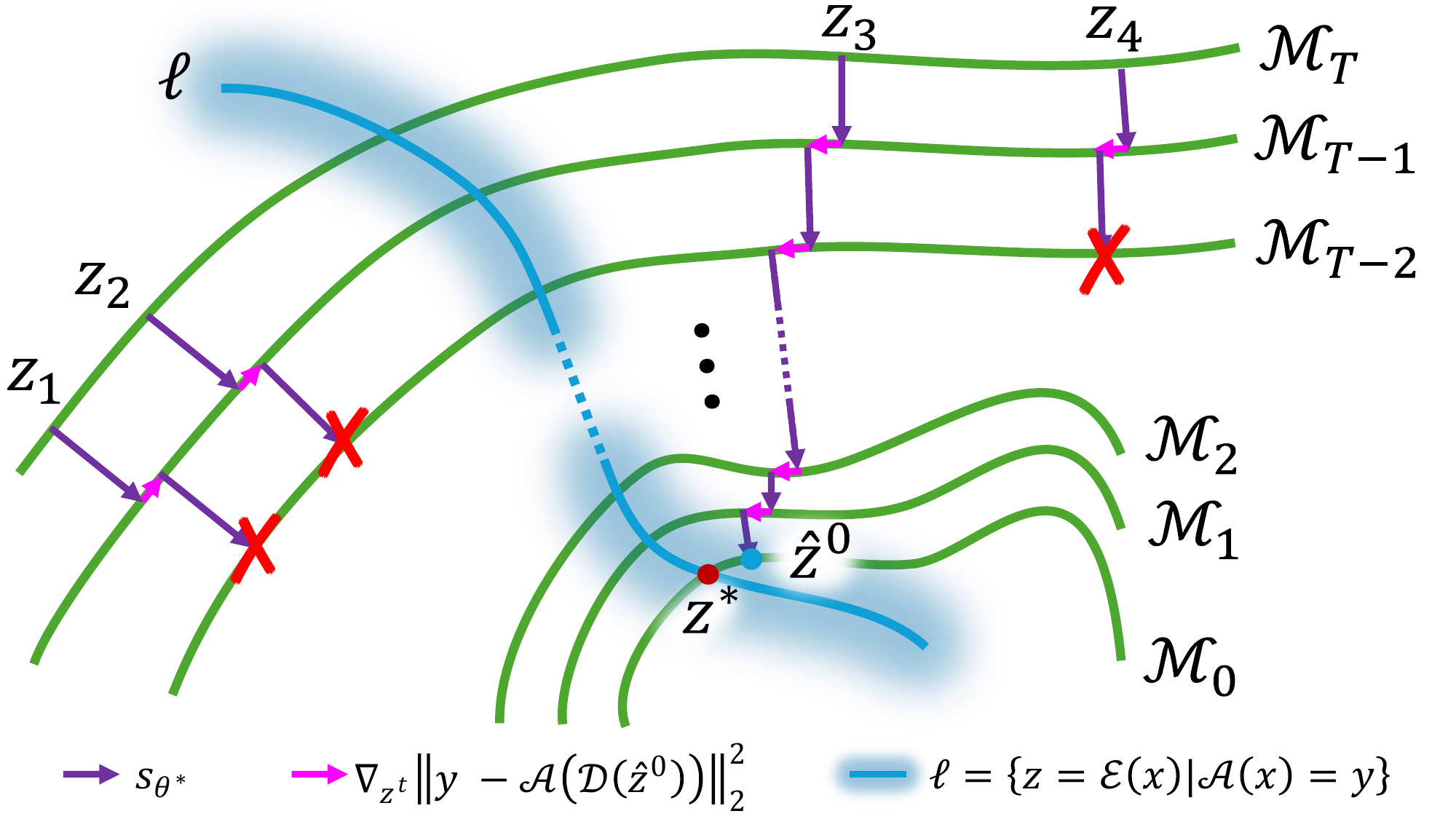}
    \caption{In the proposed PFLD method, multiple random samples, $z_l$, in the latent space of a diffusion model act as particles. 
    Starting from an outer manifold, $\mathcal{M}_T$, indistinguishable from the space of Gaussian noise, these particles progress through steps of reverse diffusion (via $s_{\theta^*}$) towards $\mathcal{M}_0$, a manifold corresponding to noise-free images. The line $\ell$ corresponds to the set of plausible solutions, all of which map to the measurement, $y$, through the forward process $\mathcal{A}$. Particles are also guided towards $\ell$ through a gradient term, $\nabla_{x^t}||y-\mathcal{A}(\mathcal{D}(\hat{z}^0_l))||_2^2$, based on intermediate estimates, $\hat{z}^0_l$, of where the particle will be at the end of reverse diffusion. Particles farther from $\ell$, destined to be far from the ground truth solution $z^*$ at the end of reverse diffusion, are pruned (red \xmark). A final singular particle reaches $\mathcal{M}_0$ producing the final estimated solution, $\hat{z}^0$.}
    \label{fig:graphical_abstract}
\end{figure}

\begin{figure}[t]
    \centering
    \includegraphics[width=1\linewidth]{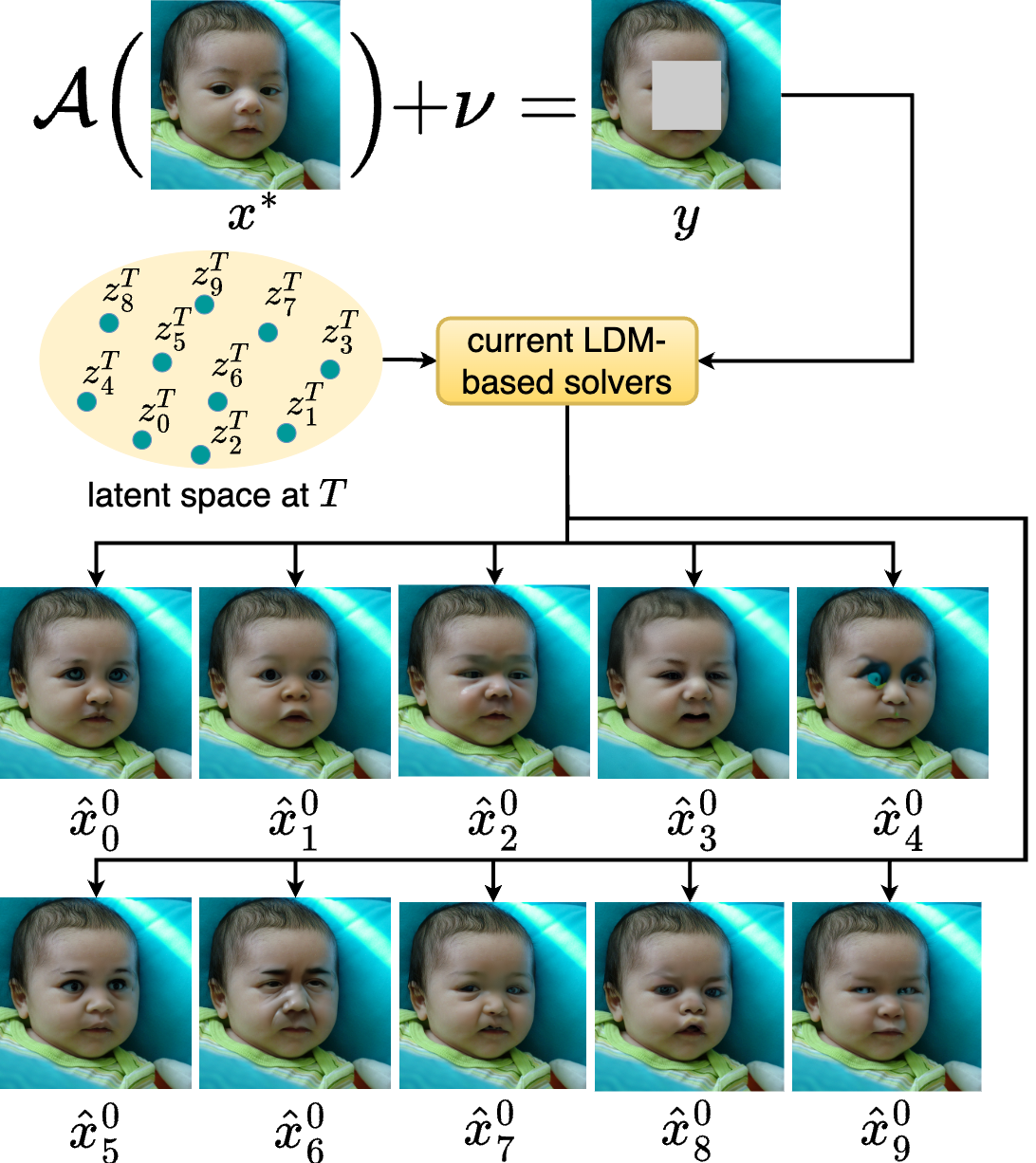}
    \caption{Ten different runs of PSLD~\cite{rout2023solving} with different initial points, $z^T_l$, produce different estimates, $\hat{x}^0_l$, of the ground truth image $x^*$. Here, $T$ is the last time step of forward path. The forward model, $\mathcal{A}$, masks the center of $x^*$. The estimator only has access to the measurement $y$ to solve the inverse problem. The image used in this figure is from the FFHQ dataset~\cite{karras2019style}.}
    \label{fig:psld_10_results}
\end{figure}

In our work, particles, $z_l$, are samples in a latent space of a diffusion model and they relate to individual solution estimates, $\hat{x}_l$, for the given inverse problem. 
Using the latent diffusion greatly improves computational efficiency compared to a pixel-space diffusion.
The latent-diffusion baseline considered here uses one Gaussian sample to initialize the reverse SDE and recover a noise-free sample, $z^0$ (or $x^0$ in pixel space). An unfavorable initialization can therefore lead to a poor reconstruction.  

Recently, filtering-based frameworks for diffusion models using multiple samples have emerged~\cite{cardoso2023monte,dou2024diffusion, wu2024practical} to reduce the issue of sensitivity arising from using only one sample. 
The method FPS-SMC~\cite{dou2024diffusion} 
requires a forward process for the given corrupted image (the measurement) and is limited to solving only \textit{linear} inverse problems. 
The Twisted Diffusion Sampler (TDS) of~\cite{wu2024practical} utilizes a Sequential Monte Carlo (SMC) technique on conditional diffusion models, and in~\cite{dou2024diffusion}, a different proposal kernel is used on SMC. 
In their reported experiments, these filtering-based methods were not applied to PSLD-style latent-diffusion inverse solvers.

Working with LDMs and high-dimensional data influenced our algorithm design. The distinction between PFLD and conceptually comparable methods is summarized in \Cref{tab:methods_diff}. The listed methods other than PFLD were not evaluated at $512 \times 512$ resolution in their original reports.

\begin{table}[ht]
    \centering
    \resizebox{\columnwidth}{!}{
    \begin{tabular}{lcccc}
    \toprule
        Method & Inverse Problem & Likelihood  & \shortstack{Particle \\ Management} & \shortstack{Measurement\\ Modification}  \\
    \midrule
        PFLD (ours) & Linear/Nonlinear & Cauchy-inspired & Pruning \& Resampling  & No \\ 
            \midrule
        FPS-SMC~\cite{dou2024diffusion} & Linear & Gaussian & Resampling  & Yes \\ 
            \midrule
        MCGdiff~\cite{cardoso2023monte} & Linear & Gaussian mixture & Resampling  & Yes \\ 
        \midrule
        TDS~\cite{wu2024practical} & Linear/Nonlinear & \shortstack{ Gaussian \& \\ Delta function} & Resampling  & No \\ 
    \bottomrule
    \end{tabular}}
    \caption{Summary of PFLD vs.\ conceptually comparable methods.}
\label{tab:methods_diff}
\end{table}

Our proposed method, PFLD, samples the latent space \textit{many} times, through the initialization of many particles.
During the process of reverse diffusion, low-scoring~/ poorly performing particles, likely not to lead to a compelling estimate, are pruned away, such that the number of remaining particles decreases and, on average, tend reliably towards a more optimal and visually compelling final solution. 
A graphical interpretation is shown in \Cref{fig:graphical_abstract}, wherein several particles, $z_l$, move towards the ground truth solution, $z^*$.
An added benefit of particle pruning is that our method is more efficient than existing SMC methods or simple repeated application of existing latent diffusion methods.
This method substantially improves the solution \textit{robustness}, such that estimates, $\hat{x}^0$, are consistently close to the ground truth, $x^*$.

The main contribution of this paper is the creation of a general particle-filtering-based framework for solving linear or nonlinear inverse problems with diffusion models. This framework can be applied to any diffusion model to improve robustness, yielding reliably improved estimates. This framework is tested upon an existing latent diffusion model, PSLD~\cite{rout2023solving}, for tasks of super-resolution, Gaussian deblurring and inpainting.

\section{Preliminaries}
\label{sec:prelim}

Diffusion models are a type of generative model that are trained to reverse a diffusion process~\cite{song2020score, deja2022analyzing, cao2024survey}. 
The diffusion process involves transforming a sample $x^0 \sim p_0(x)$ through a stochastic process defined by the Stochastic Differential Equation (SDE)
\begin{equation*}
    dx=f(x,t)\, dt+g(t)\, dw,
\end{equation*}
where $x^t$ is the input $x$ at time $t$.
By reversing the diffusion process, diffusion models~\cite{sohl2015deep, song2020improved, croitoru2023diffusion} are able to transform random samples into noise-free generated outputs.

In the formulation of pixel-space generative diffusion models, an input, $x^0$, is corrupted according to the SDE, which iteratively augments $x^0$ with noise, such that finally at $t=T$, $x^T$ is indistinguishable from a sample of a Gaussian distribution (\ie for a set of many input samples $\{x^0 \}$, the distribution of $\{x^T \}$ is approximately Gaussian).
In contrast to pixel-space diffusion models, in \emph{latent} diffusion models~\cite{rombach2022high}, the SDE and reverse SDE are applied to the \emph{latent} space representations, $z$, of a pretrained encoder, $\mathcal{E}$, and decoder, $\mathcal{D}$, pair.
The method proposed in this paper is applied to PSLD~\cite{rout2023solving}, a SOTA latent-space model.

\subsection{Utilizing Diffusion Models for Solving Inverse Problems}
\label{subsec:inverse_problem}
 
Recently, the emergence of utilizing diffusion models as generative priors~\cite{ho2020denoising, graikos2022diffusion, deja2022analyzing} for solving inverse problems has gained prominence~\cite{hu2024learning, rout2023solving, tewari2023diffusion}. In general, an inverse problem~\cite{fieguth2010statistical} can be described as the task of recovering a desired unknown and underlying quantity $x^* \in \mathbb{R}^n$ from a corrupted and noisy measurement $y \in \mathbb{R}^m$, generated via the forward process 
\begin{equation}
    y = \mathcal{A}(x^*) + \nu,
    \label{eq:general_Inverse}
\end{equation}
where $\mathcal{A}:\mathbb{R}^n \rightarrow \mathbb{R}^m$ is a deterministic corruption process and $\nu$ is usually a sample of Gaussian noise, $\nu \sim N(0,\sigma_\nu I)$.

Inverse problems and their diffusion-model-based solutions can be categorized based on whether the inverse problem is linear or non-linear, the forward model $\mathcal{A}$ is known or not, and whether the pixel or latent solution space is used.
If the forward process is \emph{known}, the transformation, $\mathcal{A}$, can be used in the solution~/ optimization process. Estimates, $\hat{x}$, are found through
\begin{equation}
    \hat{x} = \argmin_{x}{||\mathcal{A}x-y||}.
\end{equation}
If the forward process is \emph{unknown}, the problem is referred to as being \textit{blind}. Lastly, most traditional image-based inverse problems are solved in \textit{pixel} space; however, with the inclusion of an encode-decoder architecture, it is possible to solve inverse problems in the encoded \emph{latent space}.

In most inverse problems, such as inpainting, super-resolution, and deblurring, some information is \emph{lost} during the forward process, $y=\mathcal{A}(x)$. The task then of recovering the original information, $x$, from corrupted measurements, $y$, is inherently difficult due to the many-to-one nature of the problem, and $\mathcal{A}^{-1}$ does not exist. 

Recent advancements in generative modeling have led to notable progress in solving inverse problems. The framework proposed by~\cite{song2020score} utilizes score-based generative models with integrated data consistency constraints, achieving enhanced reconstruction quality and computational efficiency, though it requires extensive training data and involves complex score estimation. 
In~\cite{chung2023prompt}, Prompt-tuning Latent Diffusion Models (P2L) were introduced, which effectively address noisy and nonlinear inverse problems through posterior sampling, offering enhanced performance across various noise types but demanding significant computational resources. 
The Second-order Tweedie Sampler from Surrogate Loss (STSL)~\cite{rout2024beyond} has been developed to improve sampling quality and reduce the number of diffusion steps needed, although it requires careful parameter tuning for optimal results. 

A common theme across all latent diffusion-based methods is high sensitivity to initialization, such that results have high variability, and high computational demand, such that re-running is costly. 
Therefore, the development of a method or framework capable of producing high-quality results while reducing the computational load would be of substantial benefit and would represent a significant step towards practical usage of latent diffusion-based methods for solving inverse problems.

\subsection{Particle filtering}
In the particle filter~\cite{gordon2004beyond}, particle representations of probability density are estimated via Sequential Monte Carlo (SMC) estimation~\cite{gordon2004beyond}. 
In contrast with other linear filtering methods such as Kalman filters~\cite{welch1995introduction}, there is no need to impose the assumptions of linearity and Gaussian error covariance~\cite{fieguth2010statistical}.
This freedom makes the particle filter more broadly applicable. 

In this section, the general particle filtering algorithm is explained.
In our proposed method, a customized particle filter method is combined with the latent diffusion model. 
Therefore, the notation of diffusion models is used. 

As explained in~\cite{arulampalam2002tutorial}, the goal in the particle filter algorithm is to approximate the posterior probability density of states \(p(z^t|\{m_l\}_{l=1}^{N})\), where $z^t$ is the state and $\{m_l\}_{l=1}^{N}$ is the set of measurements, all at a given timestep $t$. 
Thus, in the particle filtering algorithm, the posterior probability density is estimated with a set of $N$ particles and their associated weights \(\{z^{t}_l,w_l^t\}_{l=1}^{N}\) as
\begin{equation}
    p(z^t|\{m_l\}_{l=1}^t) \approx      \sum_{l=1}^{N} w^t_l ~\delta(z^t-z^t_l).
    \label{eq:PDE}
\end{equation}
Here, \(\delta\) is the Dirac delta function and  \(w^t_l\) represents the weight of each particle, \(z^t_l\), in the particle filtering algorithm. 
The initial weights of all created particles are uniform, and during the process of particle filtration, the weights are subsequently adjusted according to 
\begin{equation}
    w^t_l ~~ \propto ~~w^{t-1}_l ~    \frac{p(m^t|z^t_l)~p(z^t_l|z^{t-1}_l)}{q(z^t_l|z^{t-1}_l,m^t)},
\label{eq:weights}
\end{equation}
where, \(p(m^t|z^{t}_l)\) is the likelihood function defined by the measurement model, \(q(z^t_l|z^{t-1}_l,m^t)\) is the proposal distribution function that the samples can be easily generated from, and \(p(z^t_l|z^{t-1}_l)\) is the probabilistic model of state evaluation.

In order to estimate the posterior probability density of states \(p(z^t|\{m_l\}_{l=1}^t)\) at time \(t\), we need a set of \(N\) particles and weights \(\{z^{t}_l,w_l^t\}_{l=1}^{N}\). The weights can be calculated using ~\Cref{eq:weights}. Thus, given a set of \(N\) particles \(\{z^{t-1}_l,w_l^{t-1}\}_{l=1}^{N}\) at time \(t-1\), the particle filter does the following three steps~\cite{gordon2004beyond} to estimate the new set of particles and weights:
\begin{enumerate}
    \item \textbf{Prediction}: A set of new particles are generated by sampling from \(q(z^t_l|z^{t-1}_l,m^t)\).
    \item \textbf{Updating the weights}: Given a new measurement \(m^t\), a weight \(w_l^t\) is calculated and assigned to each particle \(z_l^t\) using equation~\ref{eq:weights}. This weight \(w_l^t\) is associated with the quality of the particle \(z_l^t\) given the measurement \(m^t\).
    \item \textbf{Resampling}: 
    Typically, after several iterations, many of the particles will have a small weight and updating these particles in each time step is costly and does not tend to impact the solution. This phenomenon is called the degeneracy problem~\cite{doucet2009tutorial}. A solution to this problem is to do resampling with replacement from \(\{z_l^{t},w_l^{t}\}_{l=1}^N \) using the weights \(\{w_l^{t}\}_{l=1}^N\) that are calculated in the previous step. Resampling favours particles that have a higher quality (measured as their weight). After resampling, the weights are reset to \(w_l^{t} = \frac{1}{N}\).
\end{enumerate}

The degeneracy metric $N_d$ is related to effective sample size~\cite{salmond1990mixture} and is estimated as,
\begin{equation}
N_d \approx \frac{1}{\sum_{i=1}^{N}(w_{i}^{t})^2}.
\label{eq:degen}
\end{equation}
It can be shown that $1 \leq N_d \leq N$. The degeneracy metric is equal to $N$ after each resampling since the weights are set to $\frac{1}{N}$. A threshold $N_{th}$ is set, for which resampling is performed if ever $N_d \leq N_{th}$.

\section{Method}

\label{sec:method}

The proposed method, Particle-Filtering-Based Latent Diffusion (PFLD), integrates particle management with a diffusion-based inverse-problem solver.
We demonstrate the framework using PSLD~\cite{rout2023solving}, a strong latent-diffusion baseline whose update can be applied independently to each particle.

The algorithm for this proposed framework, integrated with PSLD, is shown in~\Cref{alg:PFLDalgorithm}. 
To be concise, the PSLD method's lines are highlighted in red in the algorithm. 
If the use of a method other than PSLD is desired, then the red lines (lines $7$ to $12$) need only be replaced with another method's diffusion process.
The particle filter essentially `wraps around' the given latent diffusion-based model, such that for each particle, reverse diffusion is carried-out independently.
The interaction between particles occurs only during resampling and pruning.
For PFLD framework, we require a set of \(N\) particles, indexed by subscript $l$, each with an associated state, $z^{t}_l$, and weight, $w_l^t$. 

In PSLD, an intermediate estimate, $\hat{z}^0_l$, of where the particle will be at the end of reverse diffusion is generated at every time step. Note that for reverse diffusion, time steps, $t$, start from $T$ and end at $0$. At each time step, the reverse SDE is applied to all particles to update their states.


In PFLD, the corrupted image, $y$, is used as the measurement at all time steps, \ie $m^t = y \  \forall t$. 
The PFLD approach fundamentally relies on its weighting process. Particle filtering algorithms often employ a Gaussian likelihood function (GLF); nevertheless, using a generic GLF in particle filtering may cause difficulties~\cite{mozhdehi2021deep}, specifically when the number of particles is limited. 
On the other hand, particle diversity is crucial in particle filtering algorithms. If particles initially are far from the true state or measurements, maintaining a reasonable weight and keeping them in the process over multiple iterations is challenging.
Because decoded particles early in the reverse SDE can differ substantially from the measurement but later become plausible reconstructions, prematurely concentrated weights are undesirable. We therefore use a Cauchy-inspired pseudo-likelihood based on the scalar measurement residual rather than asserting a normalized multivariate Cauchy density over image pixels.

The Cauchy distribution, known for its heavy tails, is more robust than the Gaussian Likelihood Function (GLF) when dealing with high-variance or outlier-prone real-world data, particularly where measurement noise follows a heavy-tailed distribution~\cite{jiang2023improving}.

\begin{figure}[t]
    \centering
    \includegraphics[width=1\linewidth]{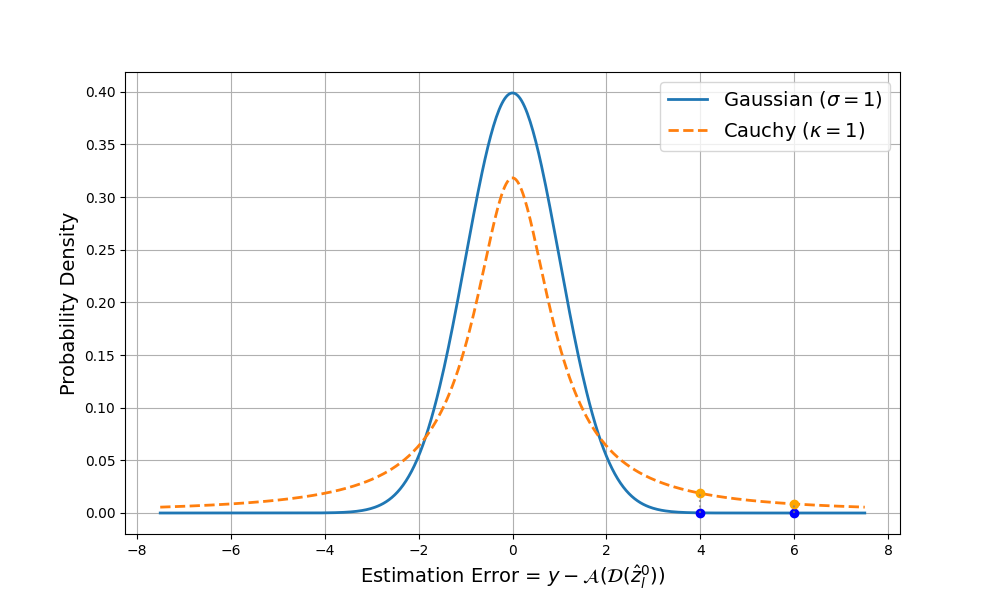}
    \caption{Comparison of Cauchy and Gaussian densities as a function of distance from a common mean at zero. The Cauchy density has a heavier tail.}
    \label{fig:cauchyvsgasussian}
\end{figure}

This choice is motivated by particle-filtering variants designed for limited particle populations~\cite{khine2023improving, jiang2023improving}. \Cref{fig:cauchyvsgasussian} compares the decay of Cauchy and Gaussian densities away from their common mean. The heavier Cauchy tail prevents the weights of particles with large early residuals from vanishing as quickly.

The scalar Cauchy density is

\begin{equation}
\operatorname{Cauchy}(x; x_0, \kappa) = \frac{1}{\pi \kappa \left[ 1 + \left( \frac{x - x_0}{\kappa} \right)^2 \right]}.
\end{equation}
For particle \(l\), define the scalar measurement residual
\begin{equation}
\rho_l = \big\|y-\mathcal{A}(\mathcal{D}(\hat{z}^0_l))\big\|_2.
\end{equation}
With \(\kappa=1\), we model the unnormalized pseudo-likelihood as
%
\begin{align}
        \widetilde{p}(y|\hat{z}^{0}_l)
        &\propto \operatorname{Cauchy}(\rho_l;0,1) \nonumber\\
        &\propto \frac{1}{\rho_l^2+1}
        = \frac{1}{\big\|y-\mathcal{A}(\mathcal{D}(\hat{z}^0_l))\big\|^2_2 + 1}.
    \label{eq:measurment_model_}
\end{align}
This pseudo-likelihood is used in the PFLD weighting function.


\subsection{Prediction} 
Generating a new set of particles starts by drawing a new particle $z^{t-1}_l$ from the proposal distribution. The proposal distribution, $q(z^{t-1}_l|z^{t}_l,y)$, is unknown and is approximated using PSLD~\cite{rout2023solving}. PSLD uses a gluing function, \(\operatorname{gluing}(\cdot)\), to encourage the gradient-guided optimization toward a consistent solution during diffusion. Thus, the proposal distribution is approximated as
\begin{align}
\label{eq:gluing} 
q(z^{t-1}_l|z^{t}_l,y) \approx z_l''^{t-1} - \operatorname{gluing}(\hat{z}^0_l,t), 
\end{align}
where $z_l''^{t-1}$ is calculated via 
\begin{equation}
    \begin{aligned}
        z_l'^{t-1} &\leftarrow \frac{\sqrt{\alpha^t}(1-\Bar{\alpha}^{t-1})}{1-\Bar{\alpha}^t} z^t_l + \frac{\sqrt{\Bar{\alpha}^{t-1}} \beta^t}{1-\Bar{\alpha}^t} \hat{z}^0_l + \Tilde{\sigma}^t \epsilon \\
        z_l''^{t-1} &\leftarrow  z_l'^{t-1} - \eta_t \nabla_{z_l^t}\big\|y - \mathcal{A}(\mathcal{D}(\hat{z}^0_l))\big\|^2_2
        \label{eq:proposal1}
    \end{aligned}
\end{equation}
as specified and explained in PSLD.
Note that $\epsilon \sim \mathcal N(0,I)$, and $\Tilde{\sigma}^t$ is provided by the trained diffusion model.
In \cref{eq:proposal1}, intermediate estimates $\hat{z}^0_l$ are directly produced using the score function $s_\theta$ of the diffusion model, through 
\begin{equation}
    \hat{z}^0_l \leftarrow \frac{1}{\sqrt{\Bar{\alpha}^t}}\big(z^t_l+(1-\Bar{\alpha}^t)  s_\theta(z_l^t,t)\big).
    \label{eq:z_hat}
\end{equation}
Producing intermediate estimates, $\hat{z}^0_l$, is the main step taken by the reverse SDE.
In~\Cref{alg:PFLDalgorithm}, Equations~\eqref{eq:gluing},~\eqref{eq:proposal1}, and~\eqref{eq:z_hat} are highlighted in red in line $7$ to $12$ as the PSLD method's core idea.



\begin{algorithm}[tb]
\caption{Proposed PFLD method}
\label{alg:PFLDalgorithm}
\textbf{Input}: $T,y,\mathcal{D},\mathcal{A},s_\theta, N, N_{th}, \operatorname{gluing}(\cdot), R$ \\
\textbf{Output}: $\mathcal{D}(\hat{z}^0)$
\begin{algorithmic}[1] 
\STATE \textit{// Generating $N$ samples from $p_T$}
\STATE $\mathcal Z \leftarrow \left\{z^{T}_l \sim \mathcal N(0,I) \middle| l \in [1, \ldots, N]\right\}$
\STATE $\mathcal W \leftarrow \left\{w^{T}_l \leftarrow \frac{1}{N} \middle| l \in [1, \ldots, N]\right\}$
\STATE \textit{// Reverse SDE}
\FOR{$t \leftarrow T-1$ to $0$}
    \FOR{$l \leftarrow 1$ to $N$}
        \vspace{-10 pt}
        \textcolor{red}{\STATE $\hat{s} \leftarrow s_\theta(z_l^t,t)$
        \STATE $\hat{z}^0_l \leftarrow \frac{1}{\sqrt{\Bar{\alpha}^t}}(z^t_l+(1-\Bar{\alpha}^t)\hat{s})$
        \STATE $\epsilon \sim \mathcal N(0,I)$
        \STATE $z_l'^{t-1} \leftarrow \frac{\sqrt{\alpha^t}(1-\Bar{\alpha}^{t-1})}{1-\Bar{\alpha}^t} z^t_l + \frac{\sqrt{\Bar{\alpha}^{t-1}} \beta^t}{1-\Bar{\alpha}^t} \hat{z}^0_l + \Tilde{\sigma}^t \epsilon$
        \STATE $z_l''^{t-1} \leftarrow z_l'^{t-1} - \eta_t \nabla_{z_l^t}\|y-\mathcal{A}(\mathcal{D}(\hat{z}^0_l))\|^2_2$
        \STATE $z_l^{t-1} \leftarrow z_l''^{t-1} - \operatorname{gluing}(\hat{z}^0_l,t)$}
        \vspace{4 pt}
        \STATE $\mathcal Z \leftarrow \{z^{t-1}_l\}$
        \STATE $w^{t-1}_l \leftarrow \frac{w^{t}_l }{\|y-\mathcal{A}(\mathcal{D}(\hat{z}^0_l))\|^2_2+1}$
        \STATE $\mathcal W \leftarrow \mathcal W \cup \{w^{t-1}_l\}$
    \ENDFOR
    
    \STATE \textit{// Normalizing the weights}
    \STATE $\mathcal W \leftarrow \left\{ w \left( \sum_{l=1}^N w_l \right)^{-1}  \middle| w \in \mathcal W\right\}$

    \STATE \textit{// Particle pruning}
    \IF{$t \mod R = 0$ \AND $N > 1$}
        \STATE \textbf{remove half of the particles with the smallest $w$ from the set of particles and weights}
        \STATE $N \leftarrow \lfloor \frac{N}{2} \rfloor$
    \ENDIF
    
    \STATE \textit{// Resampling}
    \STATE $N_d \leftarrow \frac{1}{\sum_{i=1}^{N}(w_{i}^{t})^2}$
    \IF{$N_d \leq N_{th}$}
        \FOR{$l \leftarrow 1$ to $N$}
            \STATE \textbf{draw} $j \sim \{1,\ldots,N\}$ \textbf{with probability} $\propto w_j$
            \STATE $\mathcal Z \leftarrow \left\{z^{t-1}_l \leftarrow z_j^{t-1}\right\}$
        \ENDFOR
        \STATE $ \mathcal W \leftarrow \left\{ w \leftarrow \frac{1}{N}| w \in \mathcal W \right\}$
    \ENDIF

\ENDFOR

\STATE \textbf{return} $\mathcal{D}(\hat{z}^0)$
\end{algorithmic}
\end{algorithm}


\subsection{Updating the weights} 
Our goal is for those particles likely to be near the solution at the end of reverse diffusion to remain and guide the overall particle population. 
Thus, particles should be weighted based on the closeness of $\hat{z}^0_l$, rather than the current particle state, $z^t_l$. From Equation~\eqref{eq:weights}, by replacing $m^t$ by $y$ and considering the current time index $t-1$, and the previous time index $t$ as we do a reverse SDE, we have
\begin{equation}
    w^{t-1}_l ~~ \propto ~~w^{t}_l ~    \frac{p(y|z^{t-1}_l)~p(z^{t-1}_l|z^{t}_l)}{q(z^{t-1}_l|z^{t}_l,y)}.
\label{eq:weights_sup1}
\end{equation}
Following the importance-sampling approximation in~\cite{doucet1998sequential}, we take $q(z^{t-1}_l|z^{t}_l,y) \propto p(z^{t-1}_l|z^{t}_l)$. The transition and proposal terms then cancel up to proportionality, giving
\begin{equation}
    w^{t-1}_l \propto w^{t}_l\,\widetilde{p}(y|\hat{z}^0_l),
\label{eq:weights_sup2_}
\end{equation}
and substituting \eqref{eq:measurment_model_} into \eqref{eq:weights_sup2_} yields
\begin{equation}
        w^{t-1}_l ~~ \leftarrow ~~ \frac{w^{t}_l }{\big\|y-\mathcal{A}(\mathcal{D}(\hat{z}^0_l))\big\|^2_2+1}.
        \label{eq:weight_update}
\end{equation}

\begin{table*}[tbh]
\centering
\caption{
Results for eight-fold super-resolution and Gaussian deblurring evaluated on the FFHQ-1K and ImageNet-1K datasets at $512\times512$ resolution.
We compute results only for our method (first two rows). All other results are drawn from~\cite{rout2024beyond}.
Best results are \textbf{bolded} and second best are \underline{underlined}. 
In all cases, Stable Diffusion v-1.4 and the same measurement operators as in STSL~\cite{rout2024beyond} are used. PFLD-1 is the single-particle special case corresponding operationally to the PSLD baseline; small differences from the separately reported PSLD row arise from stochastic evaluation.}

\label{tab:main_results}
\resizebox{\textwidth}{!}{%
\begin{tabular}{@{}cccccccccccccccccc@{}}
\toprule
\multirow{3}{*}{Method} &
   &
  \multicolumn{7}{c}{FFHQ-1K ($512\times512$)} &
   &
  \multicolumn{7}{c}{ImageNet-1K ($512\times512$)} &
   \\ \cmidrule(lr){3-9} \cmidrule(lr){11-17}
 &
   &
  \multicolumn{3}{c}{SR~($\times 8$)} &
   &
  \multicolumn{3}{c}{Gaussian Deblur} &
   &
  \multicolumn{3}{c}{SR~($\times 8$)} &
   &
  \multicolumn{3}{c}{Gaussian Deblur} &
   \\ \cmidrule(lr){3-5} \cmidrule(lr){7-9} \cmidrule(lr){11-13} \cmidrule(lr){15-17}
 &
   &
  LPIPS ($\downarrow$) &
  PSNR ($\uparrow$) &
  SSIM ($\uparrow$) &
   &
  LPIPS ($\downarrow$) &
  PSNR ($\uparrow$) &
  SSIM ($\uparrow$) &
   &
  LPIPS ($\downarrow$) &
  PSNR ($\uparrow$) &
  SSIM ($\uparrow$) &
   &
  LPIPS ($\downarrow$) &
  PSNR ($\uparrow$) &
  SSIM ($\uparrow$) &
   \\ \midrule

      
PFLD-10 (Ours) &
   & 0.384 & 31.52 & 89.52 &
   & \underline{0.361} & \textbf{32.59} & \underline{93.91} &
   & 0.467 & \underline{30.48} & 81.40 &
   & \underline{0.385} & \textbf{31.35} & \underline{88.77} \\ 
   
PFLD-1 (PSLD) &
   & 0.405 & 31.31 & 88.56 &
   & 0.364 & \underline{32.54} & 93.90 &
   & 0.490 & 30.23 & 79.71 &
   &0.388 & \underline{31.30} & 88.22 \\ 
   
PSLD \cite{rout2023solving} &
   & 0.402 & 31.39 & 88.89 &
   & 0.371 & 32.26 & 92.63 &
   & 0.484 & 30.23 & 80.72 &
    & 0.387 & 31.20 & 88.65 \\ \midrule

STSL \cite{rout2024beyond} &
   & \textbf{0.335} & 31.77 & 91.32 &
   & \textbf{0.308} & 32.30 & \textbf{94.04} &
   & \textbf{0.392} & \textbf{30.64} & \textbf{84.86} &
   & \textbf{0.349} & 31.03 & \textbf{90.21} \\ 

P2L &
   & 0.381 & 31.36 & 89.14 &
   & 0.382 & 31.63 & 90.89 &
   & \underline{0.441} & 30.38 & \underline{84.27} &
   & 0.402 & 30.70 & 87.92 \\ 
GML-DPS &
   & 0.368 & \underline{32.34} & \underline{92.69} &
   & 0.370 & 32.33 & 92.65 &
   & 0.482 & 30.35 & 81.01 &
    & 0.387 & 31.19 & 88.60 \\ 
LDPS &
   & \underline{0.354} & \textbf{32.54} & \textbf{93.46} &
   & 0.385 & 32.18 & 92.34 &
    & 0.480 & 30.25 & 80.87 &
   & 0.402 & 31.14 & 88.19 \\ 
DPS \cite{chung2022diffusion} &
   & 0.538 & 29.15 & 72.92 &
   & 0.694 & 28.14 & 56.15 &
   & 0.522 & 29.18 & 70.72 &
   & 0.647 & 28.11 & 55.40 \\ 
DiffPIR &
   & 0.791 & 28.12 & 42.66 &
   & 0.614 & 29.06 & 73.51 &
   & 0.740 & 28.17 & 42.80 &
   & 0.612 & 28.93 & 68.67 \\ \bottomrule
\end{tabular}%
}
\end{table*}


%
\begin{table}[!ht]
\centering
\scriptsize 
\setlength{\tabcolsep}{4pt} 
\resizebox{\columnwidth}{!}{%
\begin{tabular}{@{}c@{\hspace{2pt}}cccccccc@{}}
\toprule
\multirow{2}{*}{\shortstack{Method}} & & \multicolumn{3}{c}{ FID ($\downarrow$) - FFHQ-1K ($512\times512$)} & & \multicolumn{3}{c}{FID ($\downarrow$) - ImageNet-1K ($512\times512$)} \\ 
\cmidrule(lr){3-5} \cmidrule(lr){7-9} 
& & SR~($\times 8$) & Gaussian Deblur & Inpainting & & SR~($\times 8$) & Gaussian Deblur & Inpainting \\ \midrule
PFLD-10 (Ours) & & \textbf{16.63} & {24.35} & \textbf{16.06} & & \textbf{59.76} & \textbf{49.49} & {60.66} \\ 
PFLD-1 (PSLD) & & 18.95 & \textbf{24.12} & 16.52 & & 62.39 & {50.76} & \textbf{60.05} \\ 
\bottomrule
\end{tabular}
}
\caption{FID (InceptionV3) for PFLD-10 and PSLD (PFLD-1).
Best results are \textbf{bolded}.
}
\label{tab:fid_results}
\end{table}

\subsection{Resampling and particle pruning} 
Diffusion models require a lot of neural function evaluations (NFEs) and are computationally costly despite their exceptional performance. Consequently, the number of particles in PFLD significantly influences its running time.
In PFLD, the changes in particle states become smaller and smaller as $t\to 0$~\cite{ho2020denoising}. 
It is most advantageous to maintain a large population early in the reverse diffusion process, quickly and broadly exploring the solution space, then pruning the population down to include only the best particles, which are allowed to continue until $t=0$. The particle weights, $w^{t}_l$, act as the measure of particle fitness for the purpose of pruning, with larger weights being favored. 

In particle filtering algorithms, resampling is a solution for improving the algorithm's efficiency. Resampling is performed any time that the degeneracy metric, $N_d$, calculated via \cref{eq:degen}, falls below the resampling threshold, $N_{th}$. 
In PFLD, carefully choosing pruning schedules will enhance the effectiveness of resampling, limiting its usage to the exploration phase. 
Reducing the particle population size not only implements the exploration and exploitation idea of PFLD but also increases the method's speed (making PFLD advantageous \vs. simply running an existing LDM-based estimator multiple times to achieve a similar result).

Particle pruning is performed every $R$ steps of the reverse diffusion process and at $3R$ steps, only the single particle with largest weight will continue the reverse diffusion process. 
A na\"ive implementation simply reduces the population size by a factor of $2$ at each occurrence of pruning. The parameter $R$ determines the particle pruning schedule and, in this simple case, acts analogously to the \textit{half-life} of exponentially decaying quantities, utilized across science~\cite{rutherford1900radio}. 
\subsection{Empirical stability with increasing particle count}
For standard particle filters, estimator consistency concerns convergence toward the target quantity as the particle count \(N\) tends to infinity. Consistency results have been established for several filtering-based diffusion frameworks~\cite{cardoso2023monte,dou2024diffusion,wu2024practical}. Those results do not directly apply to PFLD because its Cauchy-inspired pseudo-likelihood and deterministic pruning alter the standard SMC construction.

We therefore make no formal consistency claim. Instead, \Cref{fig:ablation1} reports a finite-sample empirical stability study: over the tested particle counts, the reconstruction metrics tend to improve and their trial-to-trial variance tends to decrease as \(N\) grows.

\section{Results}

\label{sec:results}
The experimental setup by which our method is evaluated is first discussed, followed by the actual experimental results. 
The experiments follow the setups of PSLD and STSL to facilitate direct comparisons.
Each experiment is conducted on an NVIDIA A100 GPU.
We choose the Stable Diffusion v-1.4 model~\cite{rombach2022high} and the same measurement operators as in STSL~\cite{rout2024beyond}.

\subsection{Experimental Setup}
\paragraph{Datasets} 
Two datasets are used for evaluation: FFHQ-1K ($512\times512$)~\cite{karras2019style} and ImageNet-1K ($512\times512$)~\cite{deng2009imagenet}. For ImageNet-1K, we use the same 1,000-image selection protocol as STSL~\cite{rout2023beyond}. FFHQ contains aligned human portraits, whereas ImageNet spans a broader range of object categories and scenes; the two datasets therefore test the method on substantially different image distributions.

\newcommand{\qresultstextwidth}{0.14}
\newcommand{\qresultssuperres}{00092}
\newcommand{\qresultsinpainting}{00041}
\newcommand{\qresultsdeblur}{00775}
\newcommand{\qresultvspace}{0.5em}
\newcommand{\qresultvspacesubtr}{1em}
\newcommand{\qresulthspace}{0.5em}
\newcommand{\qresultraiseboxoffset}{1.25}
\newcommand{\qresultraiseboxoffsett}{0.95}
\newcommand{\qresultraiseboxoffsettt}{2}

\begin{figure}[t]
    \centering
    \begin{tabular}{@{\hskip \qresulthspace}c@{\hskip \qresulthspace}c@{\hskip \qresulthspace}c@{\hskip \qresulthspace}c@{\hskip \qresulthspace}}
    & Super-resolution & Inpainting & Gaussian deblur
    \vspace{\qresultvspace}
    \\
    \raisebox{\qresultraiseboxoffset\height}{\rotatebox[origin=c]{90}{Ground truth}} &
    \begin{subfigure}[t]{\qresultstextwidth\textwidth}
        \centering
        \includegraphics[width=\textwidth]{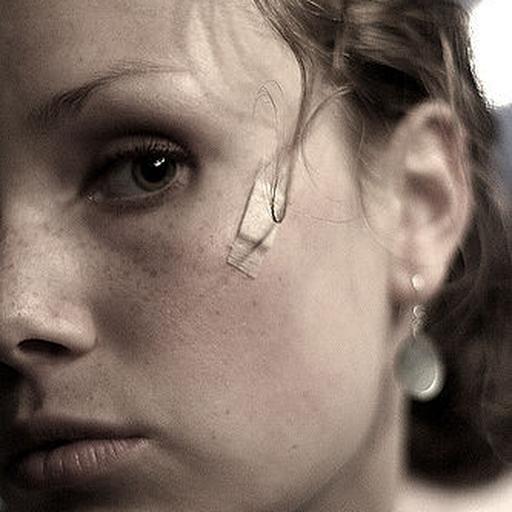}
    \end{subfigure} & 
    \begin{subfigure}[t]{\qresultstextwidth\textwidth}
        \centering
        \includegraphics[width=\textwidth]{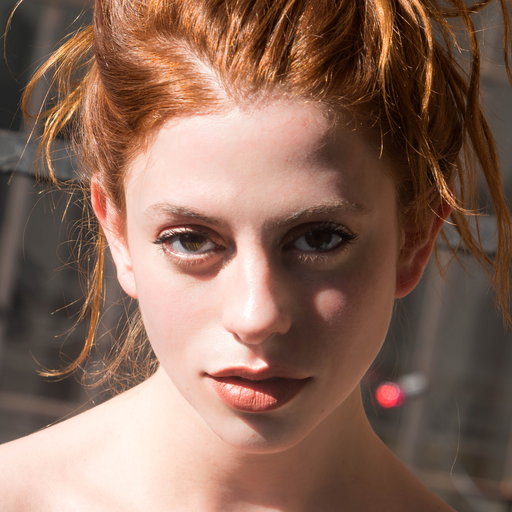}
    \end{subfigure} &
    \begin{subfigure}[t]{\qresultstextwidth\textwidth}
        \centering
        \includegraphics[width=\textwidth]{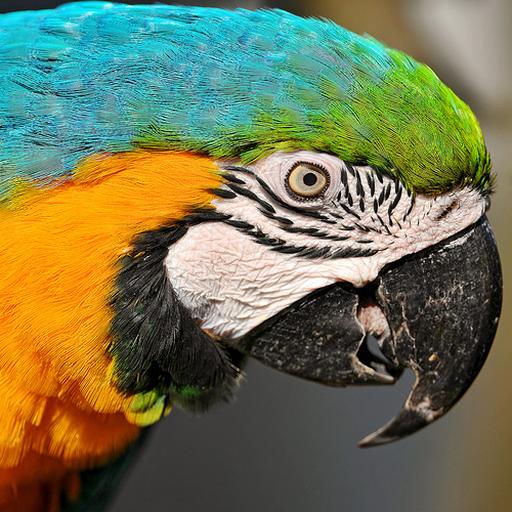}
    \end{subfigure}
    \vspace{-\qresultvspacesubtr}
    \\
%
    \raisebox{\qresultraiseboxoffsett\height}{\rotatebox[origin=c]{90}{Corrupted image}} &
    \begin{subfigure}[t]{\qresultstextwidth\textwidth}
        \centering
        \includegraphics[width=\textwidth]{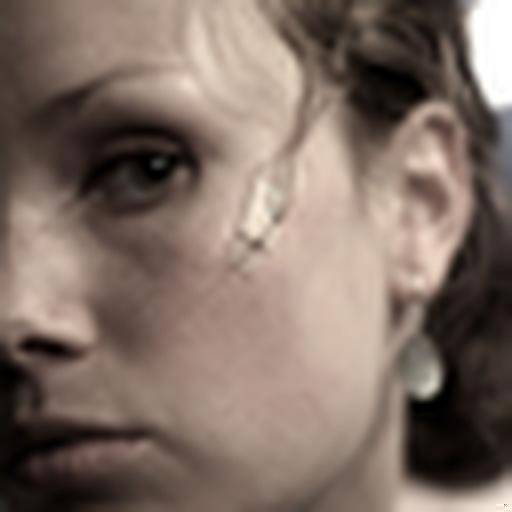}
    \end{subfigure} &
    \begin{subfigure}[t]{\qresultstextwidth\textwidth}
        \centering
        \includegraphics[width=\textwidth]{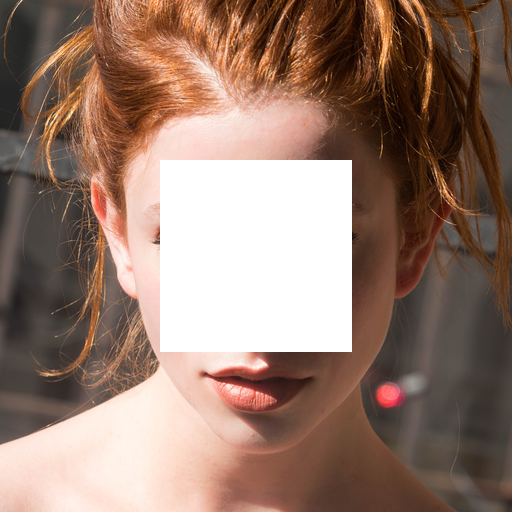}
    \end{subfigure} &
    \begin{subfigure}[t]{\qresultstextwidth\textwidth}
        \centering
        \includegraphics[width=\textwidth]{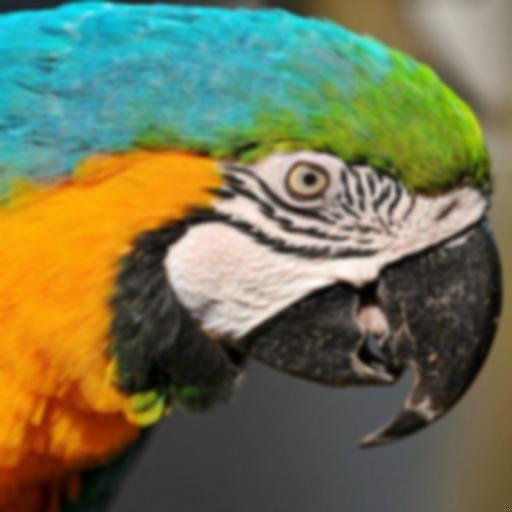}
    \end{subfigure}
    \vspace{-\qresultvspacesubtr}
    \\
%
    \raisebox{\qresultraiseboxoffsettt\height}{\rotatebox[origin=c]{90}{PSLD}} &
    \begin{subfigure}[t]{\qresultstextwidth\textwidth}
        \centering
        \includegraphics[width=\textwidth]{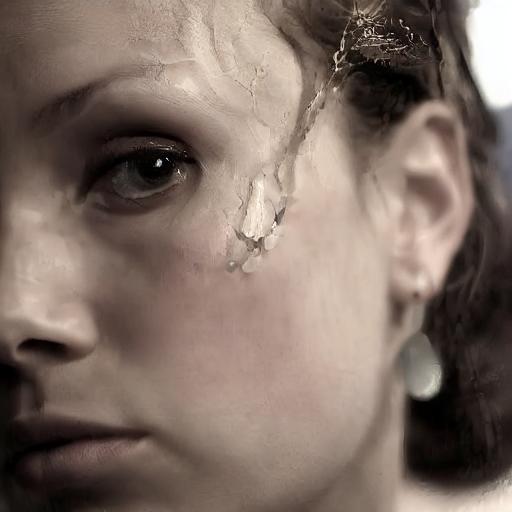}
    \end{subfigure} &
    \begin{subfigure}[t]{\qresultstextwidth\textwidth}
        \centering
        \includegraphics[width=\textwidth]{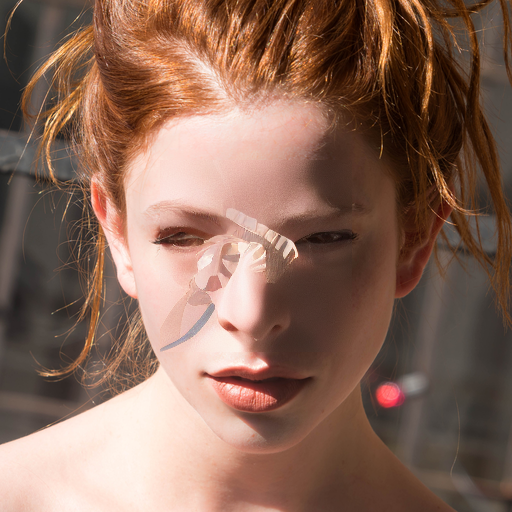}
    \end{subfigure} &
    \begin{subfigure}[t]{\qresultstextwidth\textwidth}
        \centering
        \includegraphics[width=\textwidth]{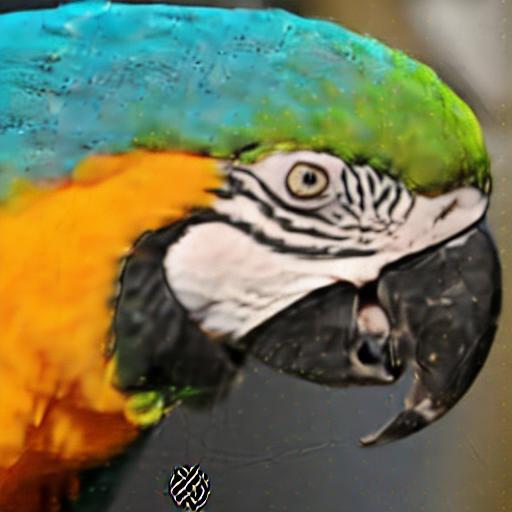}
    \end{subfigure} 
    \vspace{-\qresultvspacesubtr}
    \\
%
    \raisebox{\qresultraiseboxoffset\height}{\rotatebox[origin=c]{90}{PFLD (ours)}} &
    \begin{subfigure}[t]{\qresultstextwidth\textwidth}
        \centering
        \includegraphics[width=\textwidth]{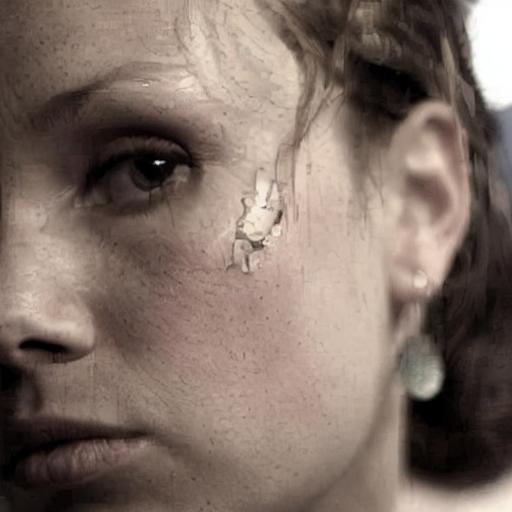}
    \end{subfigure} &
    \begin{subfigure}[t]{\qresultstextwidth\textwidth}
        \centering
        \includegraphics[width=\textwidth]{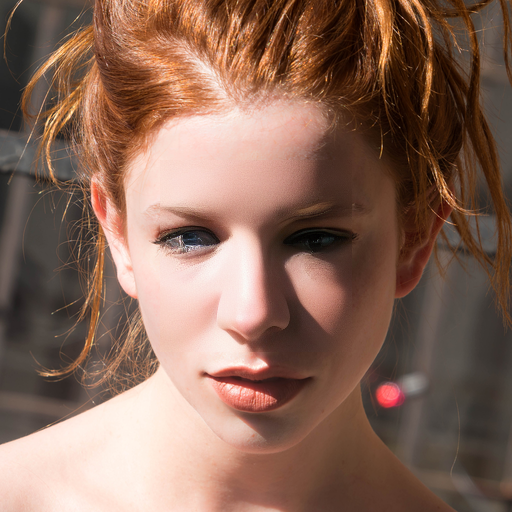}
    \end{subfigure} &
    \begin{subfigure}[t]{\qresultstextwidth\textwidth}
        \centering
        \includegraphics[width=\textwidth]{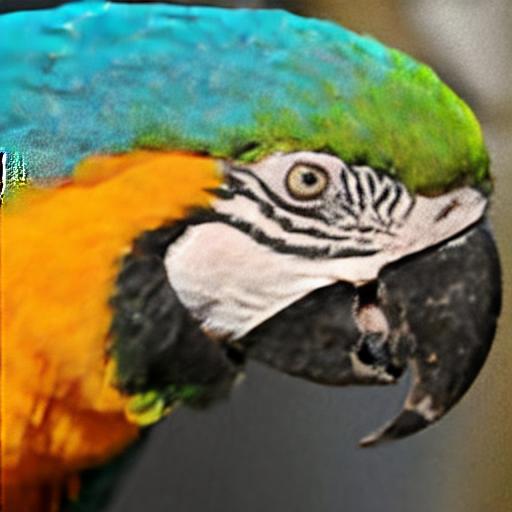}
    \end{subfigure}
    \vspace{0.1em}
    \end{tabular}
    \caption{Qualitative results. For the super-resolution task, PFLD reconstructs the face with more detail (freckles, wrinkles), while PSLD struggles to accurately reconstruct the lock of hair falling onto the bandage. For the inpainting task PSLD struggled to accurately reconstruct the eyes and the bridge of the nose, while PFLD achieved a believable result. For the Gaussian deblurring task PFLD generated a slightly sharper image with fewer artifacts. The image in the first and third columns of this figure are from ImageNet-1K ($512\times512$) and the inpainting example is from FFHQ-1K ($512\times512$).}
    \label{fig:qualitative_results}
\end{figure}

\paragraph{Baseline methods}
PSLD serves as the primary baseline, given that it is used as the foundation upon which the framework of our method, PFLD, is demonstrated in this paper.
DPS is included as a widely used pixel-space baseline. STSL~\cite{rout2023beyond} (``Beyond First-Order Tweedie'') is a strong latent-space baseline evaluated on FFHQ-1K and ImageNet-1K at $512\times512$ resolution; we compare against its reported results under the shared protocol.

\paragraph{Inverse Tasks}
As was introduced in Preliminaries, many inverse tasks within the space of image processing exist; however, for the evaluation of our method in this paper, we focus on eight-fold super-resolution, abbreviated SR~($\times 8$), Gaussian deblurring and inpainting. For super-resolution and Gaussian deblurring, we follow the same setup as STSL~\cite{rout2023beyond}, whereas for inpainting, we consider a square mask with size of $192\times192$ at the center of each image. For all tasks, the noise level is set to $\sigma_\nu = 0.01$.
 
\paragraph{Metrics} Standard metrics are used to evaluate the performance of our method: Learned Perceptual Image Patch Similarity (LPIPS)~\cite{zhang2018unreasonable}, peak signal-to-noise ratio (PSNR)~\cite{gonzalez2009digital} and structural similarity index measure (SSIM)~\cite{wang2004image}. In addition to the mentioned metrics, the Fr\'echet inception distance (FID)~\cite{heusel2017gans} is used to compare PSLD and PFLD. PSNR measures the quality of image reconstruction and does not consider local relationships between pixels in the image, hence we also report SSIM, which is sensitive to the local structure. LPIPS leverages neural network activations (in this work we use VGGNet~\cite{simonyan2015deepconvolutionalnetworkslargescale} weights) to evaluate perceptual similarity between a ground truth and a reconstructed image. Because metrics do not fully capture the visual quality of images, we also provide resultant images for qualitative visual inspection.

\begin{figure*}[t!]
    \centering
    \begin{tabular}{ccc}
        LPIPS ($\downarrow$) & PSNR ($\uparrow$) & SSIM ($\uparrow$) \\
        \begin{subfigure}{0.3\textwidth}
            \centering
            \includegraphics[width=\textwidth,trim={0.8cm 0cm 1cm 1.5cm},clip]{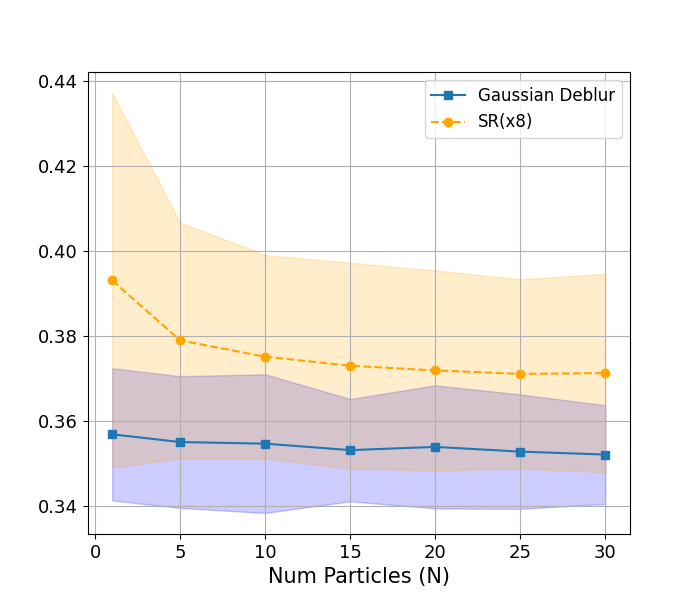}
        \end{subfigure} &
        \begin{subfigure}{0.3\textwidth}
            \centering
            \includegraphics[width=\textwidth,trim={0.8cm 0cm 1cm 1.5cm},clip]{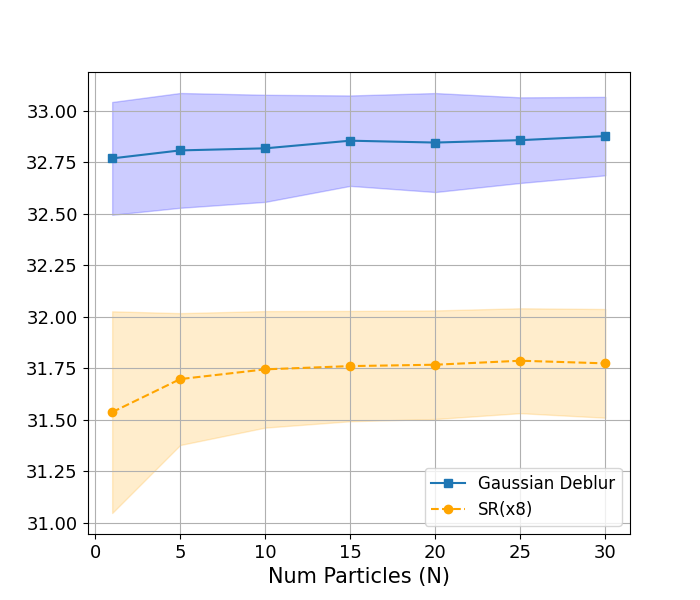}
        \end{subfigure} &
        \begin{subfigure}{0.3\textwidth}
            \centering
            \includegraphics[width=\textwidth,trim={0.8cm 0cm 1cm 1.5cm},clip]{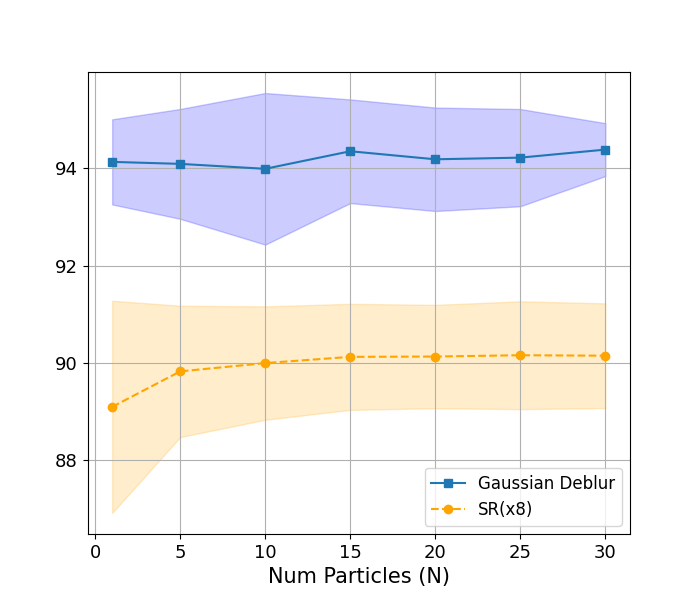}
        \end{subfigure} 
    \end{tabular}
    \caption{Effect of \(N\in\{1,5,10,15,20,25,30\}\) on reconstruction metrics for the first 100 FFHQ-1K images at $512\times512$ resolution, evaluated on eight-fold super-resolution and Gaussian deblurring. Each experiment was repeated 10 times. Shaded regions show the standard deviation across repetitions. Across the tested settings, larger particle populations generally improve the metrics and reduce trial-to-trial variability, with a stronger effect for super-resolution.}
    \label{fig:ablation1}
\end{figure*}

\paragraph{PFLD hyperparameters} We use $N=10$, based on the empirical tradeoff in \Cref{fig:ablation1}. We set the particle-pruning interval to $R=20$ to balance runtime and reconstruction quality, as shown in \Cref{tab:ablation2_results}. The resampling threshold $N_{th}$ is set to $\frac{N}{2}$ following~\cite{doucet2001sequential}. 
\subsection{Experimental Results}

We report quantitative results for PFLD in Tables~\ref{tab:main_results},~\ref{tab:fid_results}, and~\ref{tab:main_results_inpainting}, and qualitative samples in \Cref{fig:qualitative_results}. Additional examples appear in the supplementary material. For super-resolution, PFLD-10 (\ie PFLD with 10 particles) more faithfully reconstructs some facial details and text in the selected examples. For inpainting, PFLD-10 produces fewer visible artifacts in several face images. For Gaussian deblurring, most PSLD and PFLD-10 outputs are comparable in perceptual quality.


Table~\ref{tab:main_results} summarizes SR~($\times 8$) and Gaussian deblurring results for not only PFLD and PSLD, but also the reported results of several other baseline methods. 
Our method, PFLD, is primarily compared to the PSLD method. Note that PFLD with only \emph{one} particle should perform the same as PSLD, barring the effects of random initialization. Indeed, Table~\ref{tab:main_results} shows that PFLD-1 replicates reported PSLD results on both datasets for the tasks of super-resolution and Gaussian deblurring. 
PFLD-10 improves upon \mbox{PFLD-1} (and PSLD) for all metrics. 
Improvements on Gaussian deblurring are small because of artifacts, visible in Figure~\ref{fig:qualitative_results}, which are inherent to PSLD and therefore affect results from PFLD.

STSL did not report FID or evaluate inpainting. We report FID in \Cref{tab:fid_results}; PFLD-10 improves over PFLD-1 in four of the six task--dataset configurations.

Table~\ref{tab:main_results_inpainting} shows the comparison between PFLD-10 and PSLD on the inpainting task on both datasets. 
PFLD-10 outperforms PSLD on FFHQ-1K ($512\times512$) for all reported inpainting metrics, but has mixed results on the more diverse ImageNet-1K dataset.
In particular, PFLD scores marginally lower on PSNR and SSIM for this dataset.
These mixed results show that the weighting and pruning strategy does not uniformly improve every metric or dataset. In particular, measurement consistency alone may not rank perceptually preferable inpainting trajectories when the masked region is weakly constrained.

\begin{figure}[h!]
    \centering
    \includegraphics[width=0.47\textwidth]{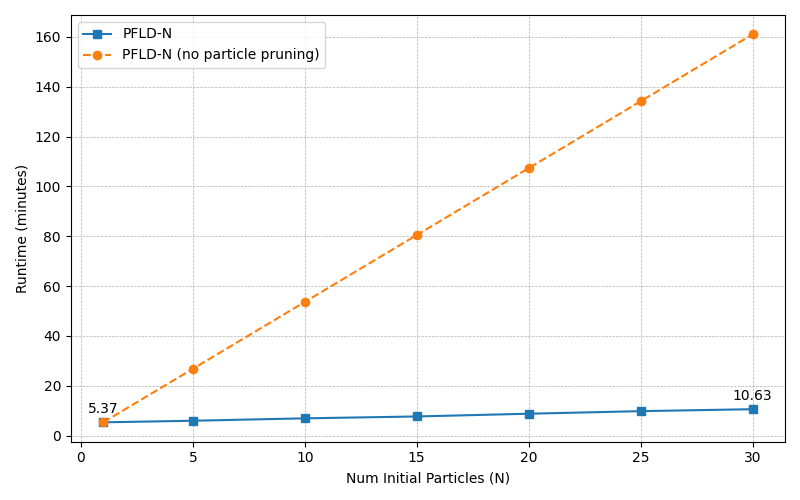}
    \caption{The effect of varying the number of initial particles on the running time of the PFLD algorithm. 
    Results are shown for the SR~($\times 8$) inverse problem for a single image from the FFHQ-1K ($512\times512$) dataset. 
    The comparison shows PFLD-N with particle pruning is far more efficient than PFLD-N (no particle pruning), with PFLD-30's running time of $10.63$~min nearly equal to PFLD-2 (no particle pruning), given that the runtime for PFLD-N (no particle pruning) is linear with the number of particles, $N$.
    As illustrated, having many initial particles and then pruning allows the proposed PFLD method to broadly explore the solution space while still being efficient.}
    \label{fig:timeplot}
\end{figure}

Figure~\ref{fig:timeplot} shows how the number of initial particles affects PFLD runtime and compares pruning against retaining every particle, whose cost is similar to running PSLD independently the same number of times.
The concept largely motivating PFLD is the hypothesis that running PFLD with a given number of initial particles will achieve similar results as selecting the best result generated from running PSLD an equivalent number of times, but that PFLD will take far less time.
The no-pruning runtime grows much more quickly with the initial population; at \(N=30\), its growth rate is approximately \(16\times\) that observed with the \(R=20\) pruning schedule.
For the tested particle counts, pruning therefore provides a substantially more efficient way to evaluate multiple initial trajectories.

\begin{table}[t]
\centering
\caption{Results of running PFLD-10 and PSLD (PFLD-1) on the inpainting task on FFHQ-1K ($512\times512$) and ImageNet-1K ($512\times512$) datasets. 
Stable Diffusion v-1.4 is used for this result with a fixed mask of $192 \times 192$ in the center of each image.
Best results are \textbf{bolded}.
}

\label{tab:main_results_inpainting}
\scriptsize 
\setlength{\tabcolsep}{4pt} 
\resizebox{\columnwidth}{!}{%
\begin{tabular}{@{}c@{\hspace{2pt}}cccccccc@{}}
\toprule
\multirow{2}{*}{\shortstack{Method}} & & \multicolumn{3}{c}{FFHQ-1K ($512\times512$)} & & \multicolumn{3}{c}{ImageNet-1K ($512\times512$)} \\ 
\cmidrule(lr){3-5} \cmidrule(lr){7-9} 
& & LPIPS ($\downarrow$) & PSNR ($\uparrow$) & SSIM ($\uparrow$) & & LPIPS ($\downarrow$) & PSNR ($\uparrow$) & SSIM ($\uparrow$) \\ \midrule
PFLD-10 (Ours) & & \textbf{0.0694} & \textbf{36.71} & \textbf{91.93} & & \textbf{0.0979} & 36.80 & {84.77} \\ 
PFLD-1 (PSLD)  & & 0.0707 & 36.70 & 91.65 & & 0.0982 & \textbf{36.81} & \textbf{84.78} \\ 
\bottomrule
\end{tabular}
}
\end{table}


\begin{table}[h!]
\centering
\caption{Ablation study of the PFLD pruning interval, evaluated on the first 100 FFHQ-1K images at $512\times512$ resolution.
We compare three schedules characterized by their rate of pruning: aggressive ($R=10$), medium ($R=20$), and gentle ($R=30$). 
Best results are \textbf{bolded}.
More rapid pruning results in a decrease in runtime, whereas more gradual pruning offers a prolonged ability to search the latent space and better results.
}
\label{tab:ablation2_results}
\scriptsize 
\setlength{\tabcolsep}{4pt} 
\resizebox{\columnwidth}{!}{%
\begin{tabular}{@{}c@{\hspace{2pt}}cccccccc@{}}
\toprule
\multirow{2}{*}{\shortstack{ \shortstack{ Pruning \\ Schedule ($R$)}}} & & \multicolumn{3}{c}{SR~($\times 8$)} & & \multicolumn{3}{c}{Gaussian Deblur} \\ 
\cmidrule(lr){3-5} \cmidrule(lr){7-9} 
& & LPIPS ($\downarrow$) & PSNR ($\uparrow$) & SSIM ($\uparrow$) & & LPIPS ($\downarrow$) & PSNR ($\uparrow$) & SSIM ($\uparrow$) \\ \midrule
10 & &0.3740 & 31.56 & 87.54 & & 0.3564 & 32.74 & {93.56} \\ 
20  & & \textbf{0.3714} & 31.62 & 87.75 & & 0.3559 & 32.80 & 93.61 \\ 
30  & & 0.3730 & \textbf{31.79} & \textbf{87.97} & & \textbf{0.3514} & \textbf{32.85} & \textbf{94.45} \\ 
\bottomrule
\end{tabular}
}
\end{table}

\subsubsection{Ablation Studies}
To better understand several key attributes of our method, we perform two ablation studies. 
In the first study, we determine the impact that the starting number of particles, $N$, has on the quality of resultant images, measured by both mean metrics and their standard deviation, to indicate the level of consistency of the proposed PFLD.
For this experiment, the first 100 images of FFHQ-1K ($512\times512$) are used, and tasks of SR~($\times 8$) and Gaussian deblurring are studied.
Results for this study are shown in Figure~\ref{fig:ablation1}. 
Observe that for both tasks, increasing the number of particles leads to an upward trend for PSNR and SSIM, as well as a downward trend for LPIPS, while simultaneously reducing the variance. 
The results also indicate that the largest rate of improvement comes from adding modest numbers of particles to the single-particle base-case, for which performance metrics most rapidly improve and standard deviation shrinks. 

The second ablation study explores the impacts of varying the aggressiveness of the particle pruning schedule and is analogous to an annealing schedule, popularized through its use in the simulated annealing~\cite{van1987simulated} technique for solving optimization problems. 
We compare three pruning schedules, parameterized by the number of time steps between pruning, $R=10, 20, 30$. Performance (LPIPS, PSNR, SSIM) is shown for all three schedules in \Cref{tab:ablation2_results}. More rapid pruning reduces computational cost, whereas gradual pruning allows for a more extensive search of the latent space and produces better results.

\section{Conclusion}
We presented PFLD, a particle-management framework for diffusion-based inverse-problem solvers. Using PSLD as the base solver, multiple latent trajectories improve reconstruction metrics and reduce trial-to-trial variability for the tested super-resolution and Gaussian-deblurring settings. The FFHQ inpainting results also improve, whereas ImageNet inpainting remains mixed. Progressive pruning makes the approach substantially less expensive than retaining every trajectory throughout reverse diffusion.

PFLD's Cauchy-inspired weighting is based only on measurement consistency, its pruning changes the standard SMC construction, and the experiments do not establish formal estimator consistency. Future work should study perceptually informed weights, alternative resampling and pruning rules, and theoretical conditions under which particle management preserves the target posterior.

\bibliographystyle{unsrt}
\bibliography{sn-bibliography}

\begin{IEEEbiography}[{\includegraphics[width=1in,height=1.25in,clip,keepaspectratio]{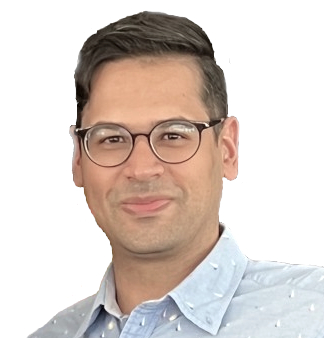}}]{Amir Nazemi} currently works as a postdoctoral researcher at the Department of Systems Design Engineering at the University of Waterloo. He has worked on many research projects with ETRI, Microsoft, and the Faculty of Health at the University of Waterloo. He received his B.Sc. and M.Sc. degrees in Computer Software Engineering and Artificial Intelligence in 2010 and 2014, respectively, and his Ph.D. in systems design engineering with a title ``Continual learning-based Video Object Segmentation'' from the University of Waterloo in Canada in 2023. His primary research interests are in continual learning and video object segmentation on long videos, computer vision, machine learning, specifically generative AI, and medical imaging.

\end{IEEEbiography}

\begin{IEEEbiography}[{\includegraphics[width=1in,height=1.25in,clip,keepaspectratio]{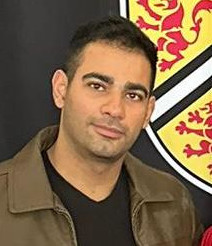}}]{M.HADI SEPANJ } is a Ph.D. candidate in the Vision and Image Processing (VIP) Lab at the University of Waterloo, Canada. He received his B.Sc. in Computer Science and M.Sc. in Artificial Intelligence before joining the Systems Design Engineering department for his doctoral studies. His research focuses on computer vision, with a particular interest in self-supervised learning, Generative Models, Statistical Machine Learning, and deep learning methods for visual understanding.
\end{IEEEbiography}

\begin{IEEEbiography}
[{\includegraphics[width=1in,height=1.25in,clip,keepaspectratio]{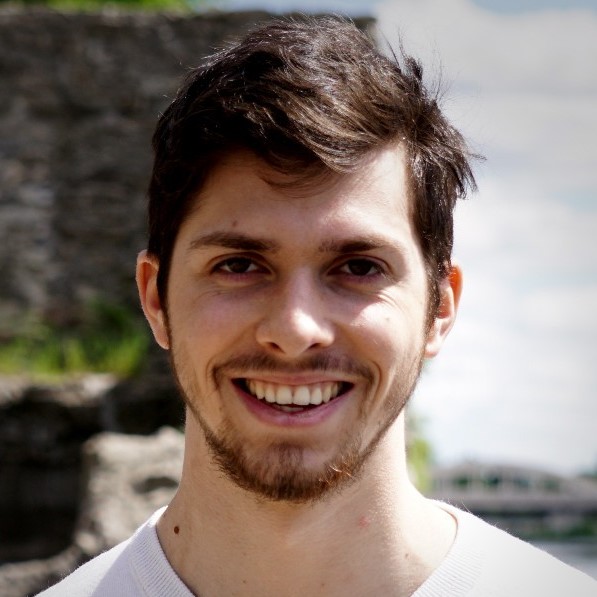}}]{Nicholas Pellegrino} is a Ph.D. candidate in the Vision and Image Processing (VIP) Lab, in the Department of Systems Design Engineering at the University of Waterloo, Canada. His BASc is in Mechatronics and MASc is in Systems Design Engineering, both from the University of Waterloo. His main research focus is on computer vision, specifically object recognition, generative models, and on methods to address label errors in training data.
\end{IEEEbiography}

\begin{IEEEbiography}[{\includegraphics[width=1in,height=1.25in,clip,keepaspectratio]{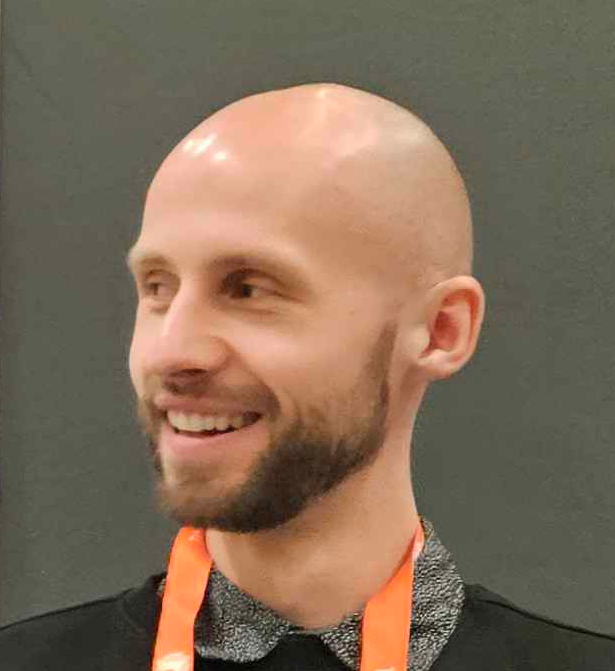}}]{Chris J. Czarnecki} is a Master's graduate from the University of Waterloo's Systems Design Engineering department, specializing in applications of machine learning in bioinformatics and computational biology as well as applications of computer vision in food computing. His current research is on microarray image denoising using transformer-based models and food portion estimation via computer vision. He is currently an AI Engineer at KisoJi Biotech, where he focuses on generative modeling for antibody development.
\end{IEEEbiography}

\begin{IEEEbiography}[{\includegraphics[width=1in,height=1.25in,clip,keepaspectratio]{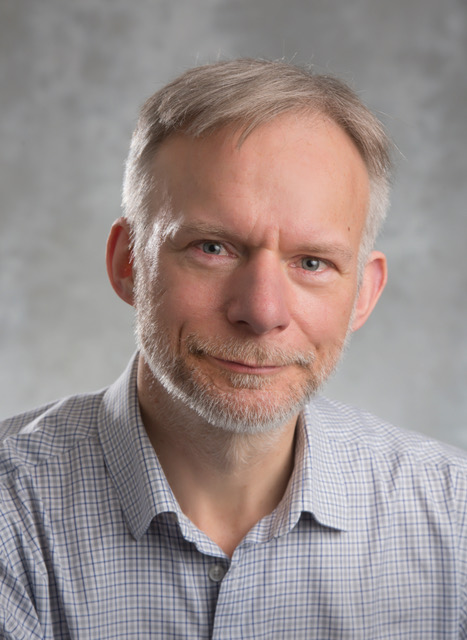}}]{Paul Fieguth} studied undergraduate Electrical Engineering at the University of Waterloo and graduate engineering degrees at the Massachusetts Institute of Technology (MIT).  He has been a member of the faculty at the University of Waterloo in Systems Design Engineering since 1996, where he has been Associate Chair Undergraduate, Department Chair, Associate Dean and, since 2023, Associate Vice President.
His research interests include statistical signal and image processing, hierarchical algorithms, data fusion, machine learning, and the interdisciplinary applications of such methods.  He has significant pedagogical interests in the area of complex systems, specifically developing a much deeper understanding among engineering students on the impact of complex systems in many areas of engineering decision making.  He is the author three textbooks, a 2010 text on Statistical Image Processing \& Multidimensional Modeling, a 2021 text on Complex Systems, and a 2022 text on Pattern Recognition.

\end{IEEEbiography}

\newpage
\makeatletter
\setlength{\@dblfptop}{0pt}
\makeatother
\section{Supplementary Materials}
\label{sec:supp}
This section contains additional findings and specific information that was not included in the main part of the paper due to space constraints.

\begin{figure*}[!t]
    \centering
    \includegraphics[width=1\linewidth]{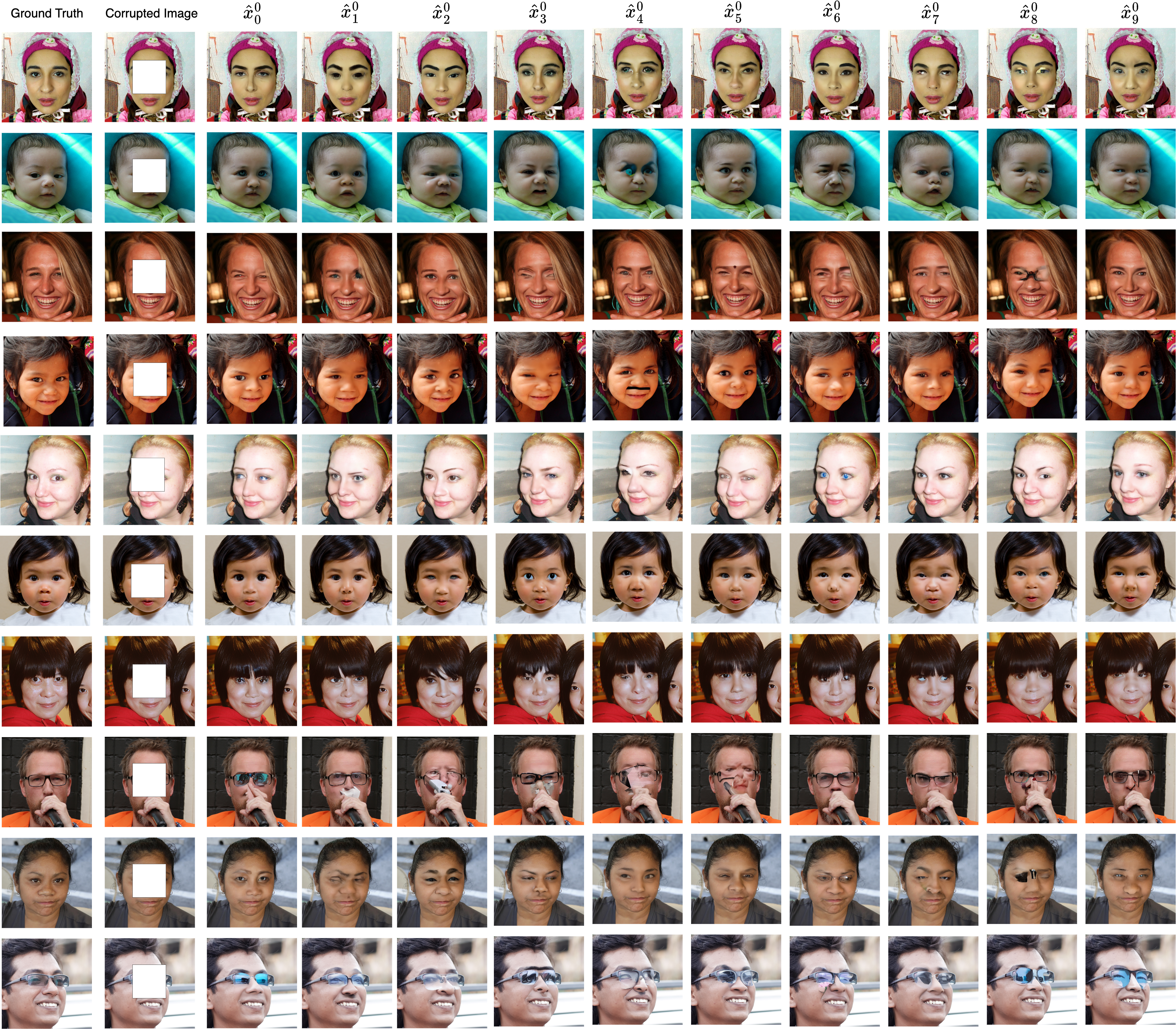}
    \caption{Ten different runs of PSLD~\cite{rout2023solving} were performed, each with a unique initialization and random seed, resulting in various outputs~$\hat{x}^0_l$. The corrupted image used for inpainting had a mask measuring $192 \times 192$ pixels, positioned in the center of the ground truth images. Images used in this figure are from the FFHQ dataset~\cite{karras2019style}. The existing diffusion-based inverse problem solvers, such as PSLD, can yield diverse outcomes when initialized differently.}
    \label{fig:psld_10_results_10}
\end{figure*}

\subsection{PSLD Results}
\label{sec:psld_results}

\Cref{fig:psld_10_results_10} shows an extension of Figure 2 of the main paper, 
showing different results of PSLD~\cite{rout2023solving} for different initializations of the PSLD algorithm.
The inconsistency between results of subsequent trials is the main problem of diffusion-based solvers and serves as the main motivation for the proposed Particle-Filtering-based Latent Diffusion (PFLD) framework.

\subsection{Results and Evaluation}

\subsubsection{Quantitative results}

Standard implementations of LPIPS, PSNR and SSIM metrics in the \textit{TorchMetrics} package were used for calculating the quantitative results.

\subsubsection{Qualitative results}
Figures~\ref{fig:qualitative_results_1}--\ref{fig:qualitative_results_3} show selected ImageNet-1K examples for eight-fold super-resolution and Gaussian deblurring, together with FFHQ-1K inpainting examples. In several cases, PFLD-10 reconstructs text or facial details more plausibly and suppresses artifacts present in the PSLD output. Other examples, particularly some inpainting and deblurring cases, show little visible difference.

\Cref{fig:qualitative_results_4} provides a counterexample in which PFLD-10 is worse than PSLD for super-resolution and comparable for the other two tasks. Figures~\ref{fig:qualitative_results_5} and~\ref{fig:qualitative_results_6} show additional FFHQ-1K super-resolution and Gaussian-deblurring examples. These selected results illustrate both the potential benefit and the limitations of ranking trajectories by measurement consistency.

\newcommand{\qresultstextwidthzero}{0.22}
\newcommand{\qresultssuperreszero}{00208}
\newcommand{\qresultsinpaintingzero}{00015}
\newcommand{\qresultsdeblurzero}{00089}

\begin{figure*}[htbp]
    \begin{tabular}{cccc}
        Ground Truth & Corrupted Image & PSLD & PFLD-10 (ours) \\
        \raisebox{1.75\height}{\rotatebox[origin=c]{90}{Super-resolution}}
    \centering
    \begin{subfigure}[b]{\qresultstextwidthzero\textwidth}
        \centering
        \includegraphics[width=\textwidth]{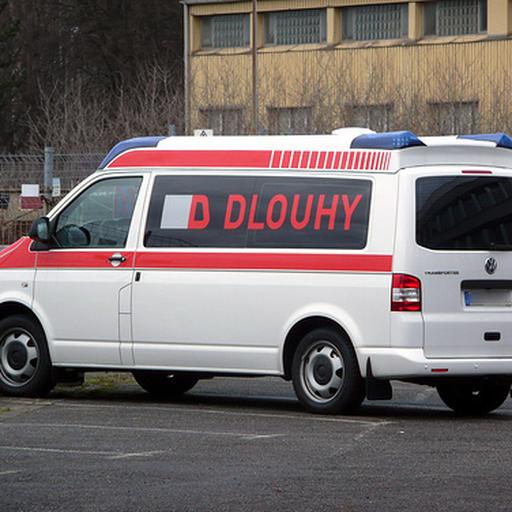}
    \end{subfigure} &
    \begin{subfigure}[b]{\qresultstextwidthzero\textwidth}
        \centering
        \includegraphics[width=\textwidth]{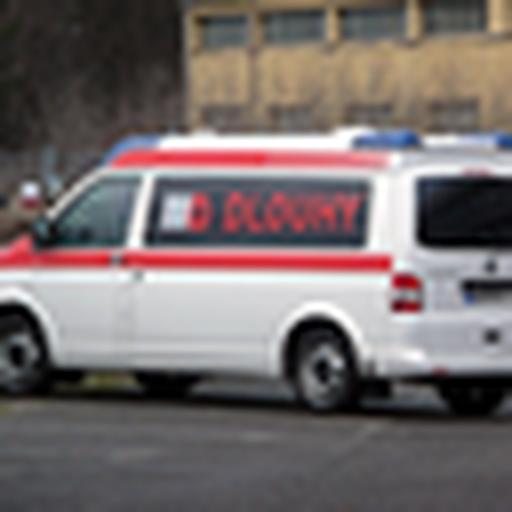}
    \end{subfigure} &
    \begin{subfigure}[b]{\qresultstextwidthzero\textwidth}
        \centering
        \includegraphics[width=\textwidth]{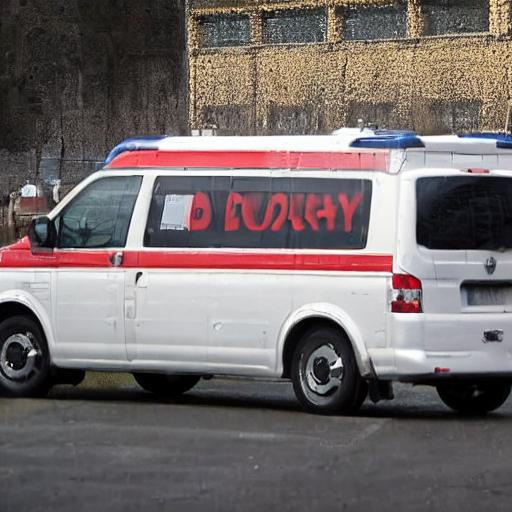}
    \end{subfigure} &
    \begin{subfigure}[b]{\qresultstextwidthzero\textwidth}
        \centering
        \includegraphics[width=\textwidth]{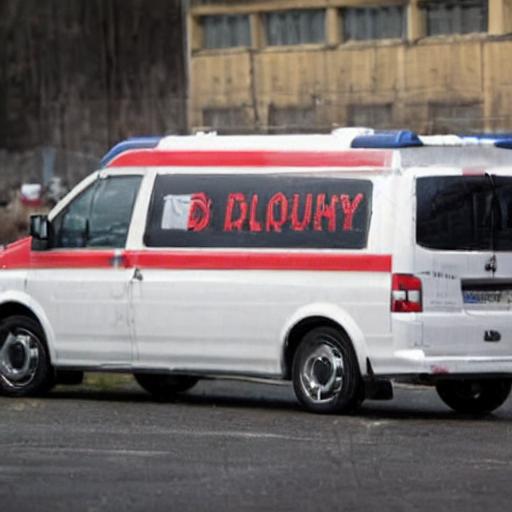}
    \end{subfigure} \\
    \raisebox{2.75\height}{\rotatebox[origin=c]{90}{Inpainting}}
    \begin{subfigure}[b]{\qresultstextwidthzero\textwidth}
        \centering
        \includegraphics[width=\textwidth]{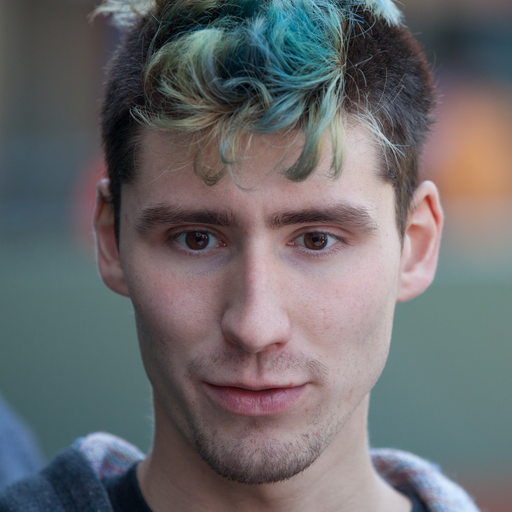}
    \end{subfigure} &
    \begin{subfigure}[b]{\qresultstextwidthzero\textwidth}
        \centering
        \includegraphics[width=\textwidth]{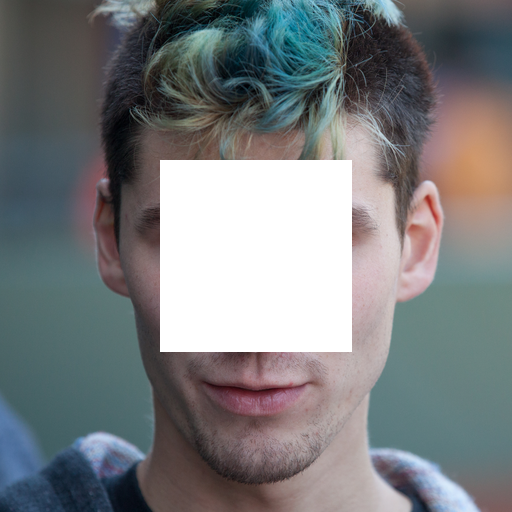}
    \end{subfigure} &
    \begin{subfigure}[b]{\qresultstextwidthzero\textwidth}
        \centering
        \includegraphics[width=\textwidth]{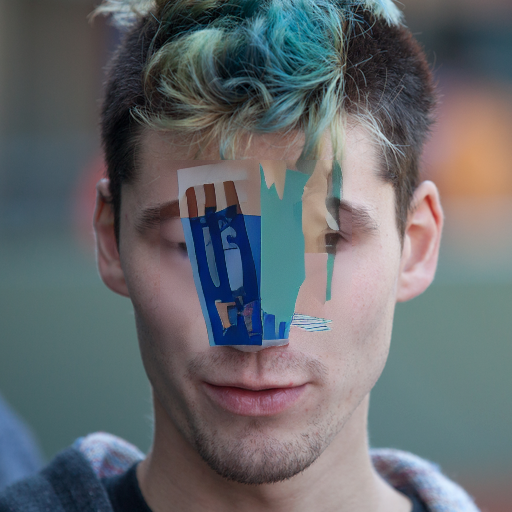}
    \end{subfigure} &
    \begin{subfigure}[b]{\qresultstextwidthzero\textwidth}
        \centering
        \includegraphics[width=\textwidth]{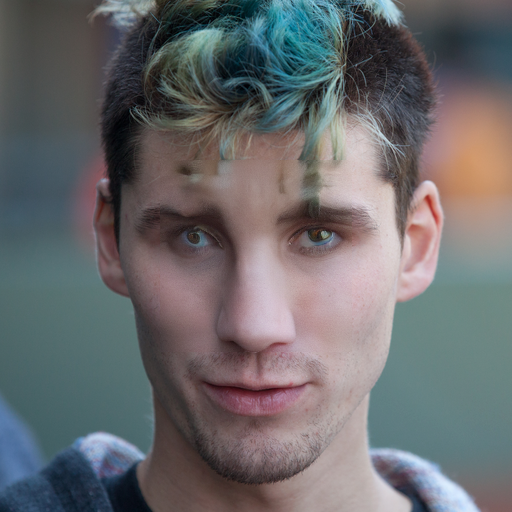}
    \end{subfigure} \\
    \raisebox{1.5\height}{\rotatebox[origin=c]{90}{Gaussian Deblur}}
    \begin{subfigure}[b]{\qresultstextwidthzero\textwidth}
        \centering
        \includegraphics[width=\textwidth]{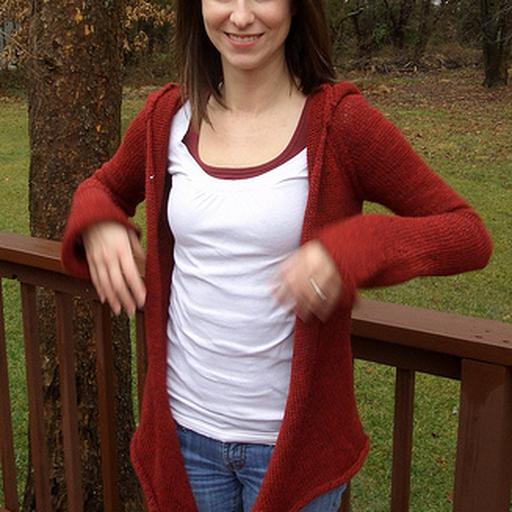}
    \end{subfigure} &
    \begin{subfigure}[b]{\qresultstextwidthzero\textwidth}
        \centering
        \includegraphics[width=\textwidth]{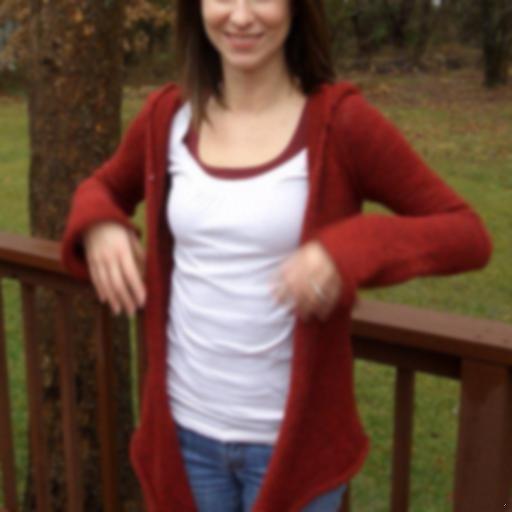}
    \end{subfigure} &
    \begin{subfigure}[b]{\qresultstextwidthzero\textwidth}
        \centering
        \includegraphics[width=\textwidth]{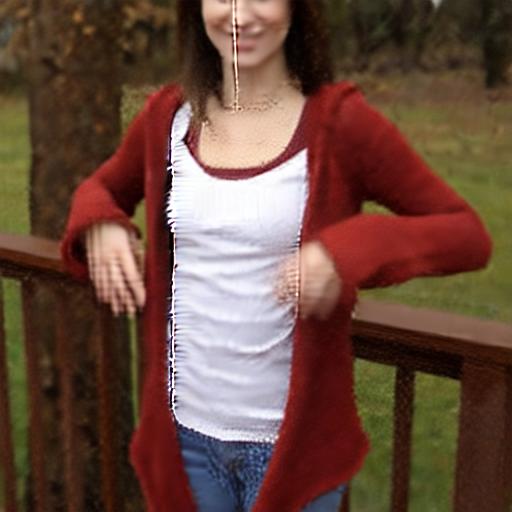}
    \end{subfigure} &
    \begin{subfigure}[b]{\qresultstextwidthzero\textwidth}
        \centering
        \includegraphics[width=\textwidth]{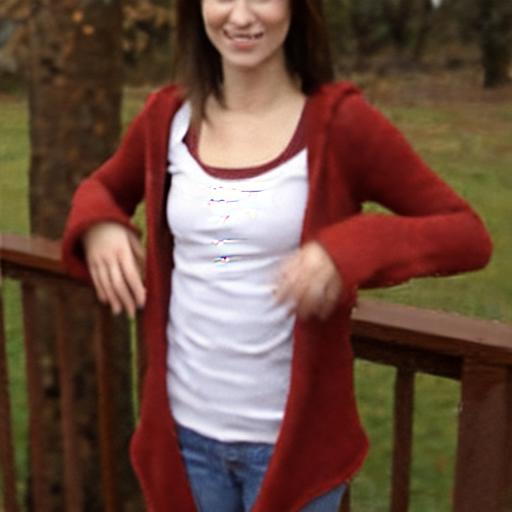}
    \end{subfigure}
    \end{tabular}
    \caption{In the super-resolution task, PFLD performed a better reconstruction of the printed text on the ambulance, whereas PSLD struggled to do so. PSLD struggled to recreate the face for the inpainting task, whereas PFLD produced an acceptable outcome. During the Gaussian Deblurring task, PFLD produced a slightly sharper image with fewer artifacts. The first and third rows of this figure show images from ImageNet-1K ($512\times512$), while the inpainting example comes from FFHQ-1K ($512\times512$).}
    \label{fig:qualitative_results_1}
\end{figure*}

\newcommand{\qresultssuperresone}{00541}
\newcommand{\qresultsinpaintingone}{00064}
\newcommand{\qresultsdeblurone}{00557}

\begin{figure*}[htbp]
    \begin{tabular}{cccc}
        Ground Truth & Corrupted Image & PSLD & PFLD-10 (ours) \\
        \raisebox{1.75\height}{\rotatebox[origin=c]{90}{Super-resolution}}
    \centering
    \begin{subfigure}[b]{\qresultstextwidthzero\textwidth}
        \centering
        \includegraphics[width=\textwidth]{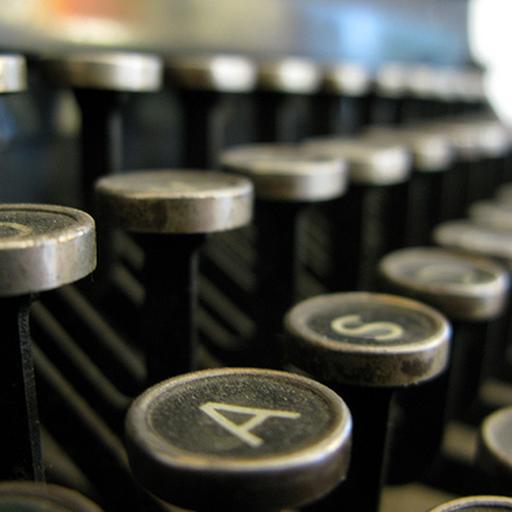}
    \end{subfigure} &
    \begin{subfigure}[b]{\qresultstextwidthzero\textwidth}
        \centering
        \includegraphics[width=\textwidth]{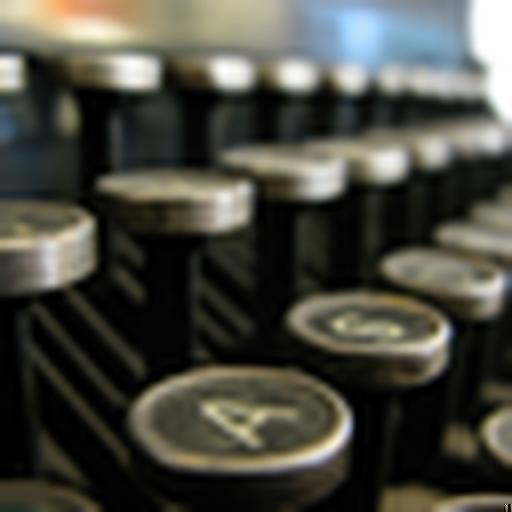}

    \end{subfigure} &
    \begin{subfigure}[b]{\qresultstextwidthzero\textwidth}
        \centering
        \includegraphics[width=\textwidth]{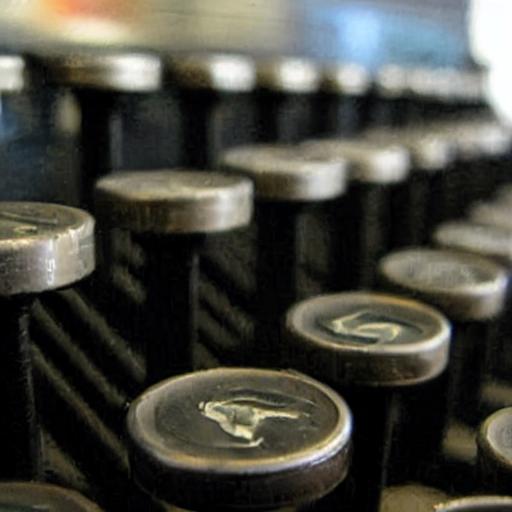}
    \end{subfigure} &
    \begin{subfigure}[b]{\qresultstextwidthzero\textwidth}
        \centering
        \includegraphics[width=\textwidth]{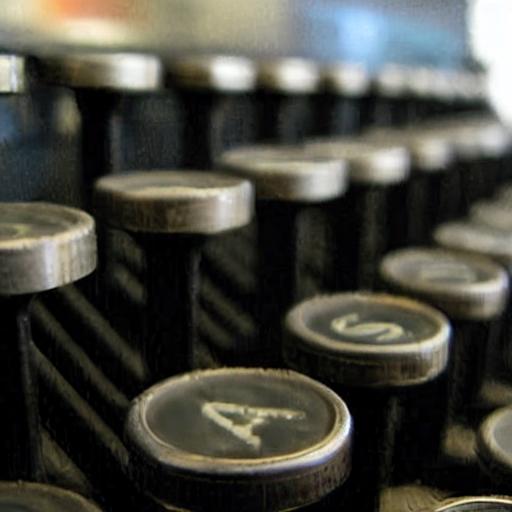}
    \end{subfigure} \\
    \raisebox{2.75\height}{\rotatebox[origin=c]{90}{Inpainting}}
    \begin{subfigure}[b]{\qresultstextwidthzero\textwidth}
        \centering
        \includegraphics[width=\textwidth]{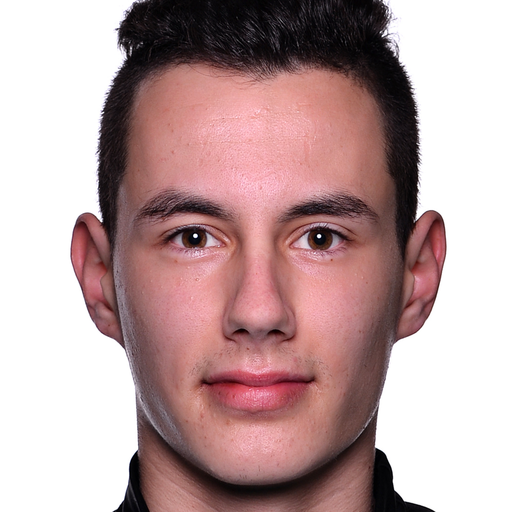}
    \end{subfigure} &
    \begin{subfigure}[b]{\qresultstextwidthzero\textwidth}
        \centering
        \includegraphics[width=\textwidth]{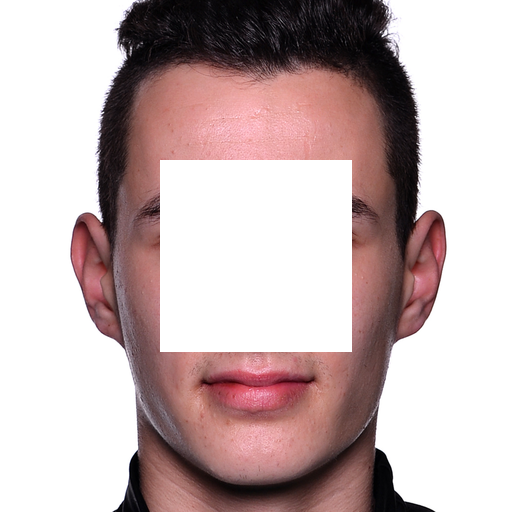}
    \end{subfigure} &
    \begin{subfigure}[b]{\qresultstextwidthzero\textwidth}
        \centering
        \includegraphics[width=\textwidth]{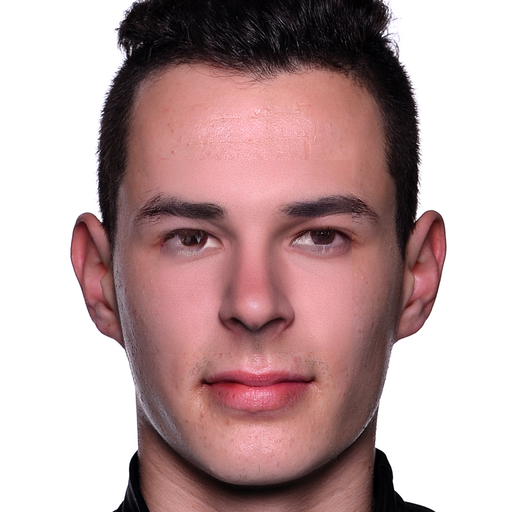}
    \end{subfigure} &
    \begin{subfigure}[b]{\qresultstextwidthzero\textwidth}
        \centering
        \includegraphics[width=\textwidth]{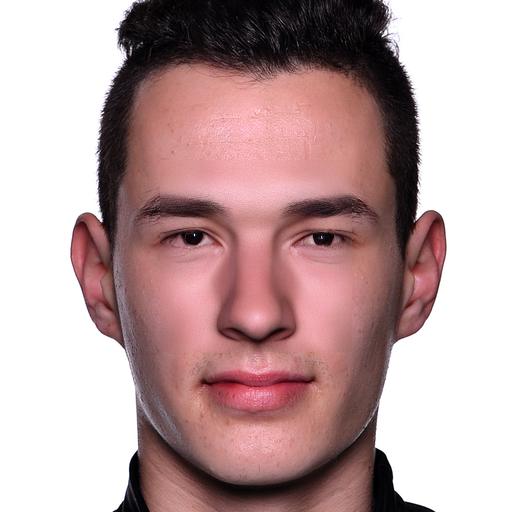}
    \end{subfigure} \\
    \raisebox{1.5\height}{\rotatebox[origin=c]{90}{Gaussian Deblur}}
    \begin{subfigure}[b]{\qresultstextwidthzero\textwidth}
        \centering
        \includegraphics[width=\textwidth]{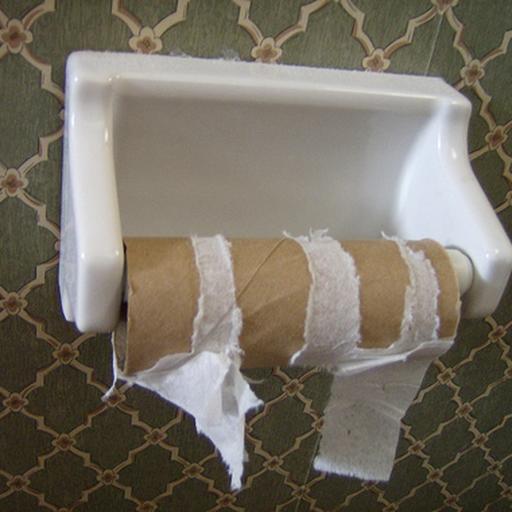}
    \end{subfigure} &
    \begin{subfigure}[b]{\qresultstextwidthzero\textwidth}
        \centering
        \includegraphics[width=\textwidth]{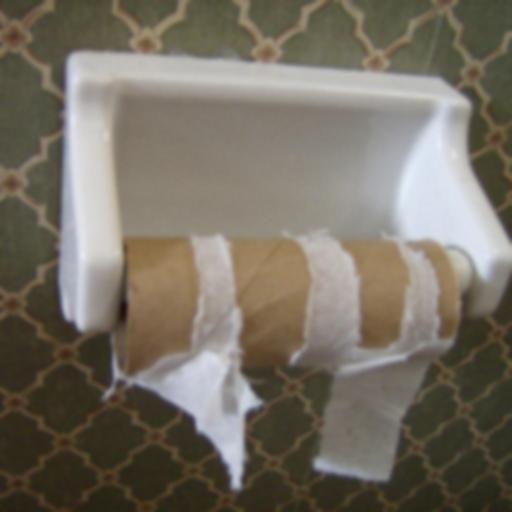}
    \end{subfigure} &
    \begin{subfigure}[b]{\qresultstextwidthzero\textwidth}
        \centering
        \includegraphics[width=\textwidth]{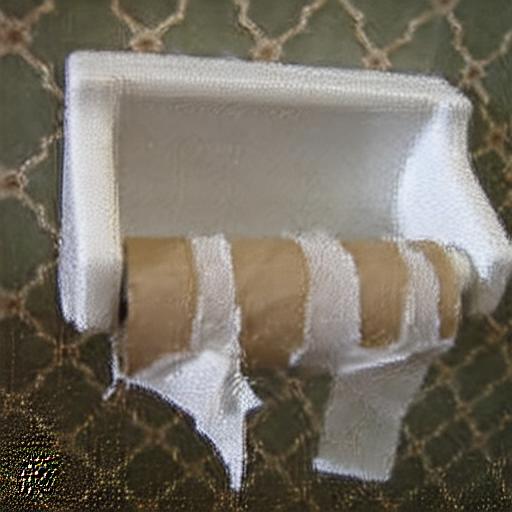}
    \end{subfigure} &
    \begin{subfigure}[b]{\qresultstextwidthzero\textwidth}
        \centering
        \includegraphics[width=\textwidth]{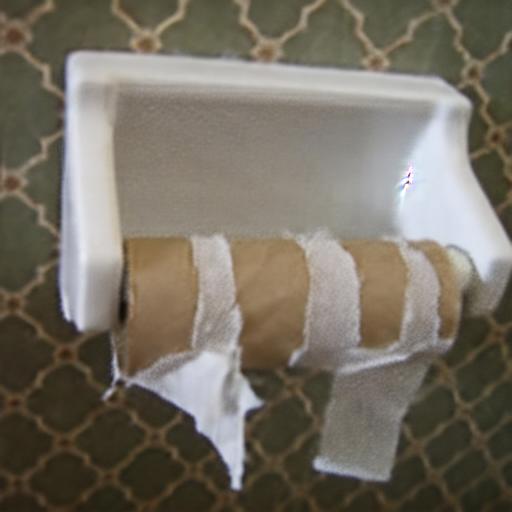}
    \end{subfigure}
    \end{tabular}
    \caption{PFLD outperformed PSLD in reconstructing characters on the typewriter keys, "A" and "S," over the super-resolution task; however, its performance in inpainting and Gaussian deblurring is comparable to PSLD. The first and third rows of this figure show images from ImageNet-1K ($512\times512$) images are used for super resolution and Gaussian deblurring and for inpainting FFHQ-1K ($512\times512$) is used.}
    \label{fig:qualitative_results_2}
\end{figure*}

\newcommand{\qresultssuperrestwo}{00324}
\newcommand{\qresultsinpaintingtwo}{00033}
\newcommand{\qresultsdeblurtwo}{00151}

\begin{figure*}[htbp]
    \begin{tabular}{cccc}
        Ground Truth & Corrupted Image & PSLD & PFLD-10 (ours) \\
        \raisebox{1.75\height}{\rotatebox[origin=c]{90}{Super-resolution}}
    \centering
    \begin{subfigure}[b]{\qresultstextwidthzero\textwidth}
        \centering
        \includegraphics[width=\textwidth]{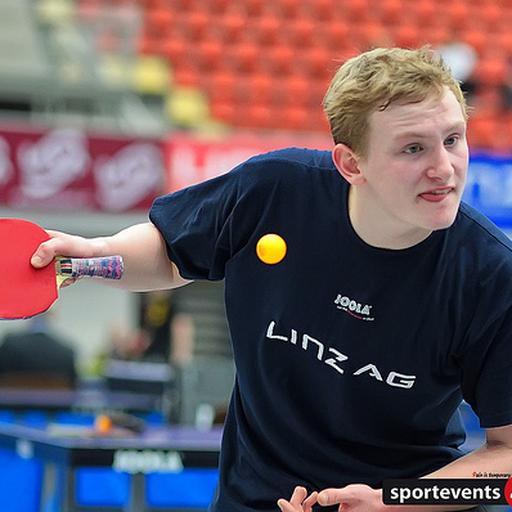}
    \end{subfigure} &
    \begin{subfigure}[b]{\qresultstextwidthzero\textwidth}
        \centering
        \includegraphics[width=\textwidth]{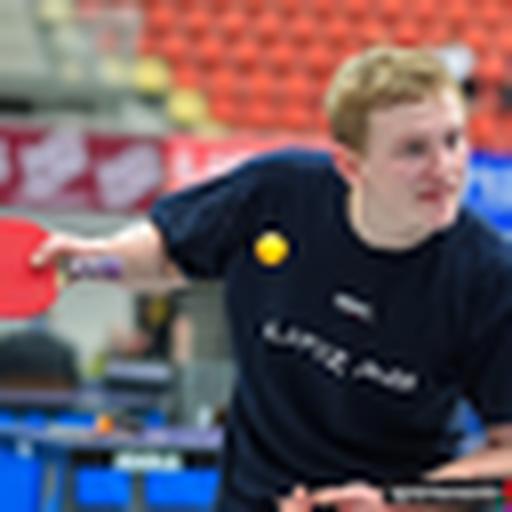}
    \end{subfigure} &
    \begin{subfigure}[b]{\qresultstextwidthzero\textwidth}
        \centering
        \includegraphics[width=\textwidth]{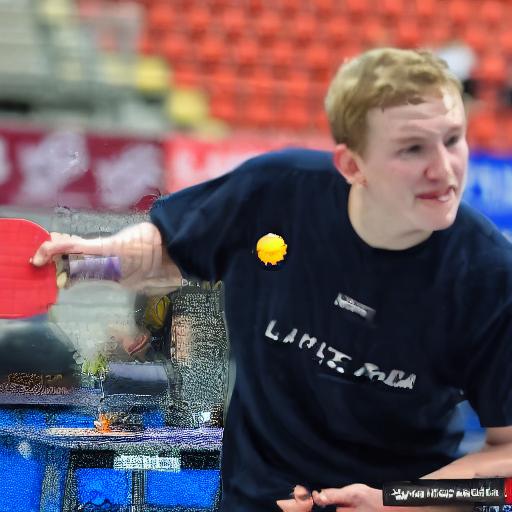}
    \end{subfigure} &
    \begin{subfigure}[b]{\qresultstextwidthzero\textwidth}
        \centering
        \includegraphics[width=\textwidth]{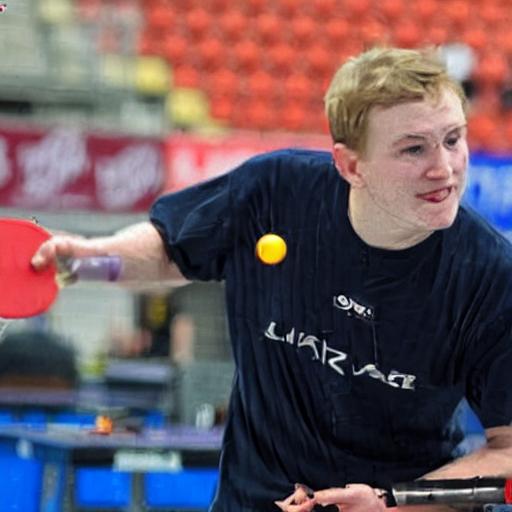}
    \end{subfigure} \\
    \raisebox{2.75\height}{\rotatebox[origin=c]{90}{Inpainting}}
    \begin{subfigure}[b]{\qresultstextwidthzero\textwidth}
        \centering
        \includegraphics[width=\textwidth]{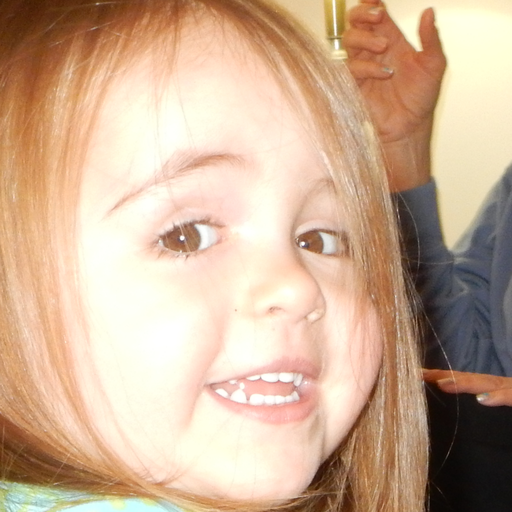}
    \end{subfigure} &
    \begin{subfigure}[b]{\qresultstextwidthzero\textwidth}
        \centering
        \includegraphics[width=\textwidth]{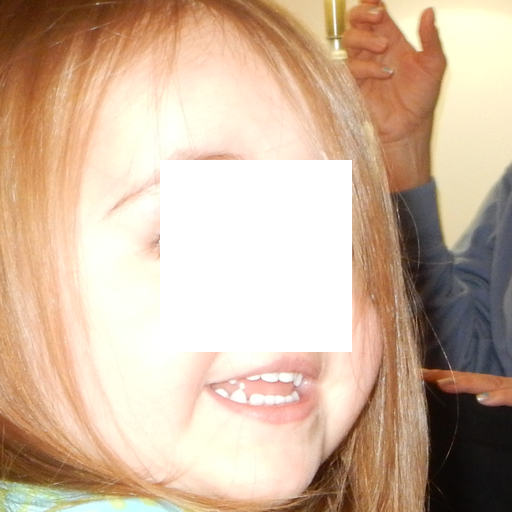}
    \end{subfigure} &
    \begin{subfigure}[b]{\qresultstextwidthzero\textwidth}
        \centering
        \includegraphics[width=\textwidth]{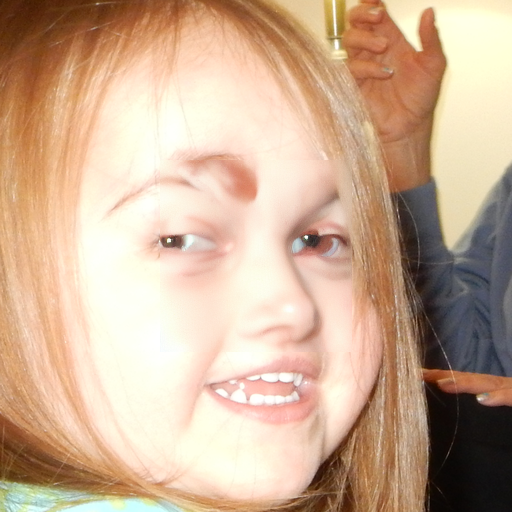}
    \end{subfigure} &
    \begin{subfigure}[b]{\qresultstextwidthzero\textwidth}
        \centering
        \includegraphics[width=\textwidth]{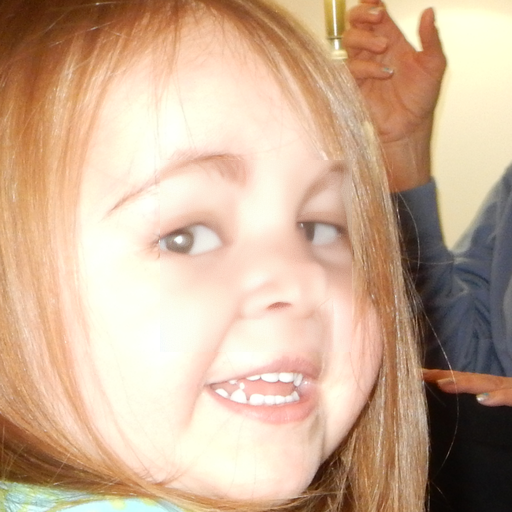}
    \end{subfigure} \\
    \raisebox{1.5\height}{\rotatebox[origin=c]{90}{Gaussian Deblur}}
    \begin{subfigure}[b]{\qresultstextwidthzero\textwidth}
        \centering
        \includegraphics[width=\textwidth]{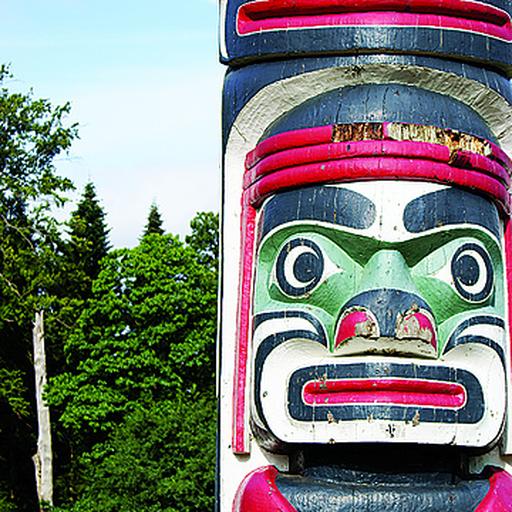}
    \end{subfigure} &
    \begin{subfigure}[b]{\qresultstextwidthzero\textwidth}
        \centering
        \includegraphics[width=\textwidth]{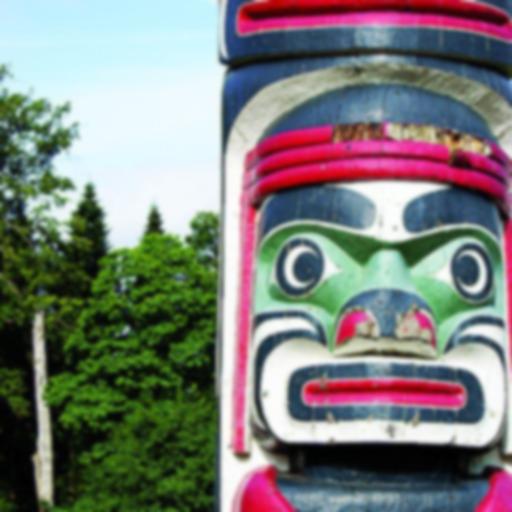}
    \end{subfigure} &
    \begin{subfigure}[b]{\qresultstextwidthzero\textwidth}
        \centering
        \includegraphics[width=\textwidth]{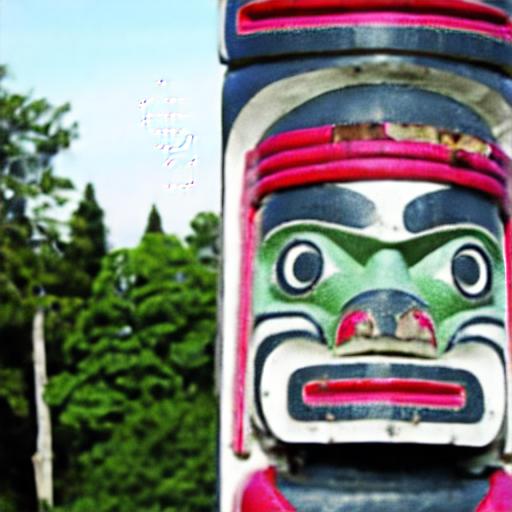}
    \end{subfigure} &
    \begin{subfigure}[b]{\qresultstextwidthzero\textwidth}
        \centering
        \includegraphics[width=\textwidth]{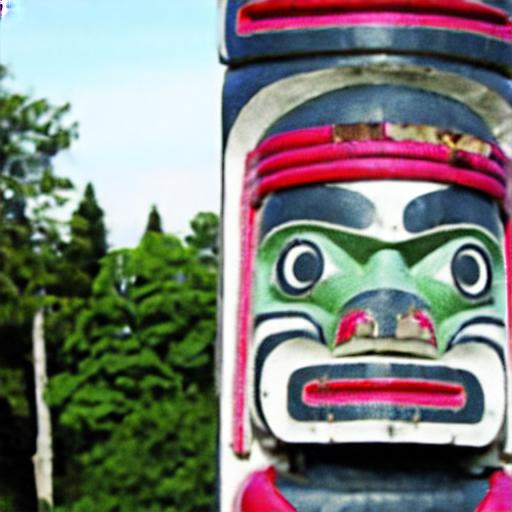}
    \end{subfigure}
    \end{tabular}
    \caption{Qualitative comparison on three selected examples. PFLD-10 reconstructs the player's face more faithfully in the super-resolution example. The inpainting outputs differ visibly, while the Gaussian-deblurring outputs have comparable overall quality and the PFLD-10 result contains fewer artifacts. The first and third rows are from ImageNet-1K; the second row is from FFHQ-1K. All images are $512\times512$.}
    \label{fig:qualitative_results_3}
\end{figure*}

\newcommand{\qresultssuperresthree}{00399}
\newcommand{\qresultsinpaintingthree}{00048}
\newcommand{\qresultsdeblurthree}{00580}

\begin{figure*}[htbp]
    \begin{tabular}{cccc}
        Ground Truth & Corrupted Image & PSLD & PFLD-10 (ours) \\
        \raisebox{1.75\height}{\rotatebox[origin=c]{90}{Super-resolution}}
    \centering
    \begin{subfigure}[b]{\qresultstextwidthzero\textwidth}
        \centering
        \includegraphics[width=\textwidth]{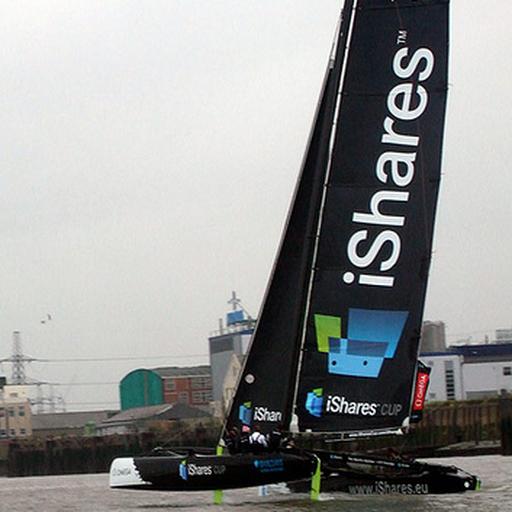}
    \end{subfigure} &
    \begin{subfigure}[b]{\qresultstextwidthzero\textwidth}
        \centering
        \includegraphics[width=\textwidth]{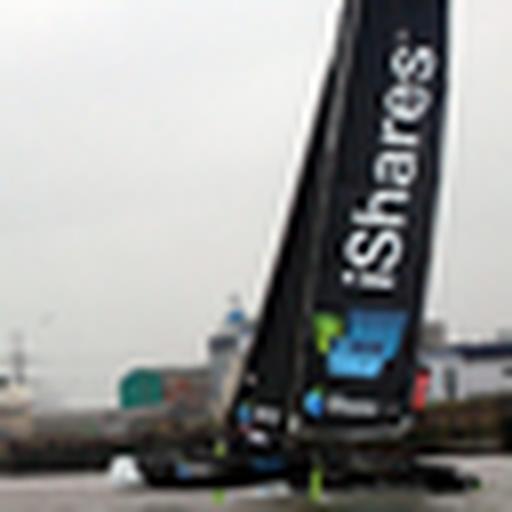}
    \end{subfigure} &
    \begin{subfigure}[b]{\qresultstextwidthzero\textwidth}
        \centering
        \includegraphics[width=\textwidth]{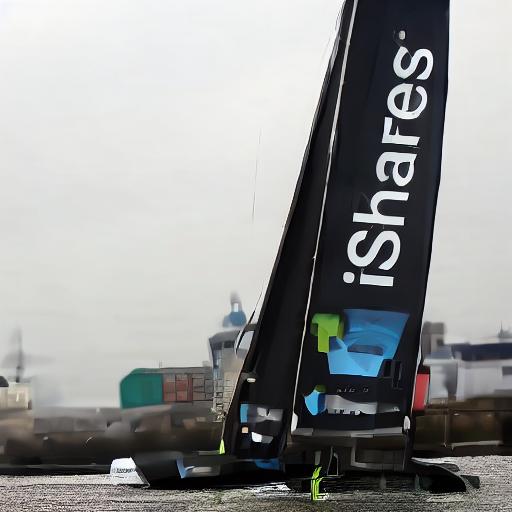}
    \end{subfigure} &
    \begin{subfigure}[b]{\qresultstextwidthzero\textwidth}
        \centering
        \includegraphics[width=\textwidth]{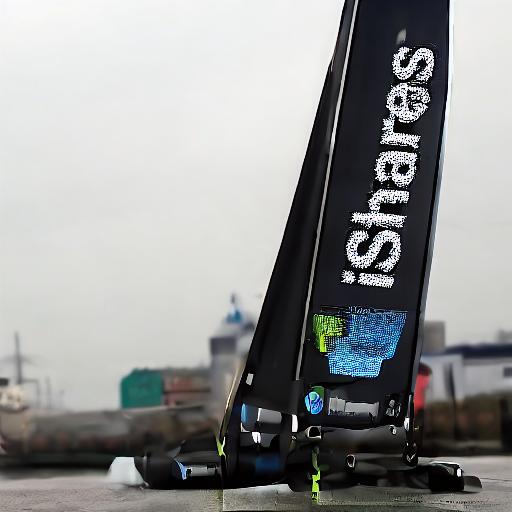}
    \end{subfigure} \\
    \raisebox{2.75\height}{\rotatebox[origin=c]{90}{Inpainting}}
    \begin{subfigure}[b]{\qresultstextwidthzero\textwidth}
        \centering
        \includegraphics[width=\textwidth]{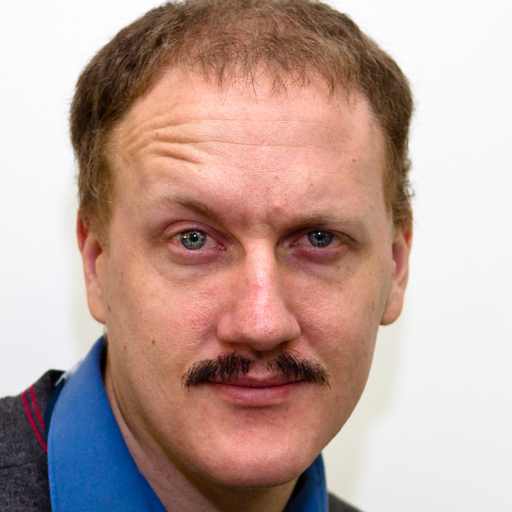}
    \end{subfigure} &
    \begin{subfigure}[b]{\qresultstextwidthzero\textwidth}
        \centering
        \includegraphics[width=\textwidth]{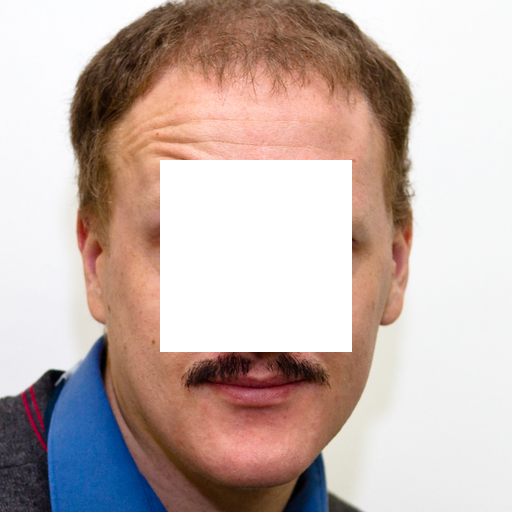}
    \end{subfigure} &
    \begin{subfigure}[b]{\qresultstextwidthzero\textwidth}
        \centering
        \includegraphics[width=\textwidth]{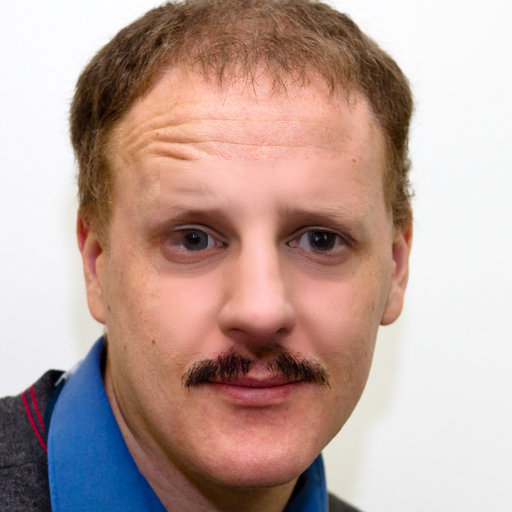}
    \end{subfigure} &
    \begin{subfigure}[b]{\qresultstextwidthzero\textwidth}
        \centering
        \includegraphics[width=\textwidth]{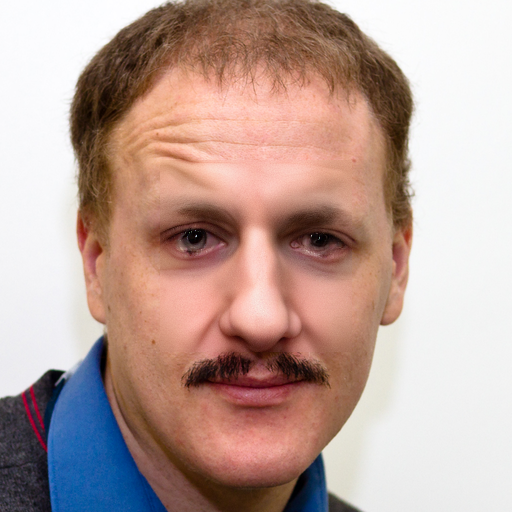}
    \end{subfigure} \\
    \raisebox{1.5\height}{\rotatebox[origin=c]{90}{Gaussian Deblur}}
    \begin{subfigure}[b]{\qresultstextwidthzero\textwidth}
        \centering
        \includegraphics[width=\textwidth]{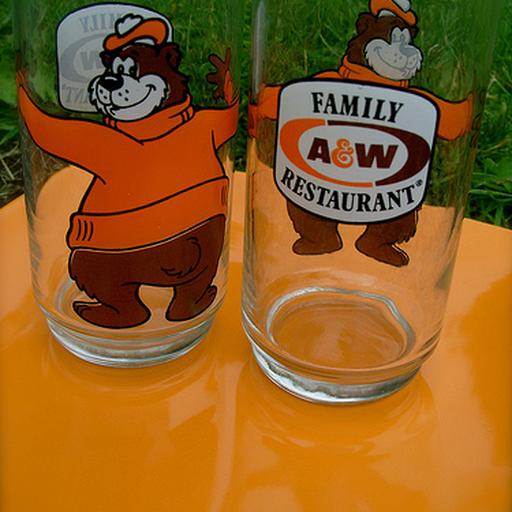}
    \end{subfigure} &
    \begin{subfigure}[b]{\qresultstextwidthzero\textwidth}
        \centering
        \includegraphics[width=\textwidth]{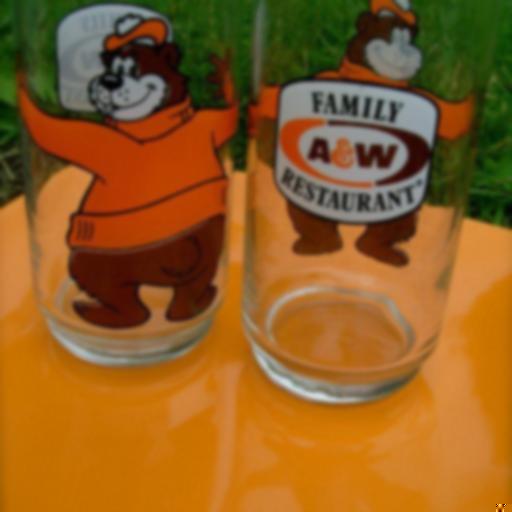}
    \end{subfigure} &
    \begin{subfigure}[b]{\qresultstextwidthzero\textwidth}
        \centering
        \includegraphics[width=\textwidth]{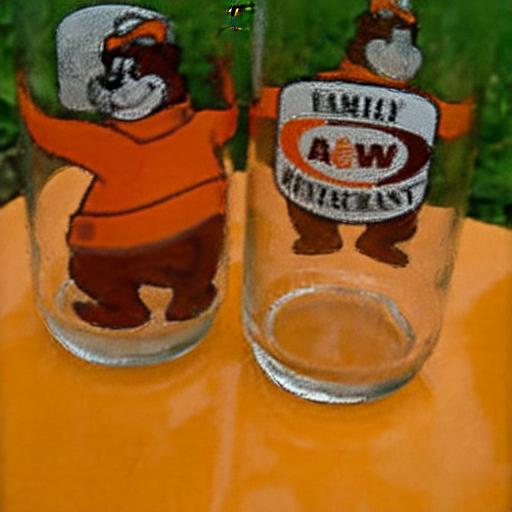}
    \end{subfigure} &
    \begin{subfigure}[b]{\qresultstextwidthzero\textwidth}
        \centering
        \includegraphics[width=\textwidth]{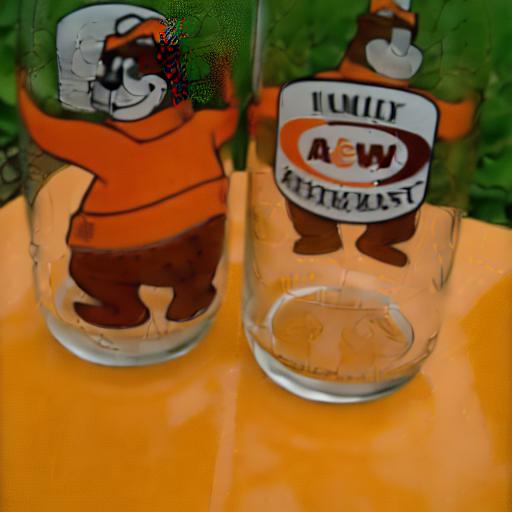}
    \end{subfigure}
    \end{tabular}
    \caption{For Inpainting and Gaussian Deblurring, PFLD offers similar outcomes as PSLD; however, for the super-resolution task, PFLD showed inferior performance compared to PSLD. Here again, first and third rows of this figure show images from ImageNet-1K ($512\times512$), while the inpainting example comes from FFHQ-1K ($512\times512$).}
    \label{fig:qualitative_results_4}
\end{figure*}

\newcommand{\qresultssuperresffhqa}{00024}
\newcommand{\qresultssuperresffhqb}{00052}
\newcommand{\qresultssuperresffhqc}{00073}

\begin{figure*}[htbp]
    \begin{tabular}{cccc}
        Ground Truth & Corrupted Image & PSLD & PFLD-10 (ours) \\
    \centering
    \begin{subfigure}[b]{\qresultstextwidthzero\textwidth}
        \centering
        \includegraphics[width=\textwidth]{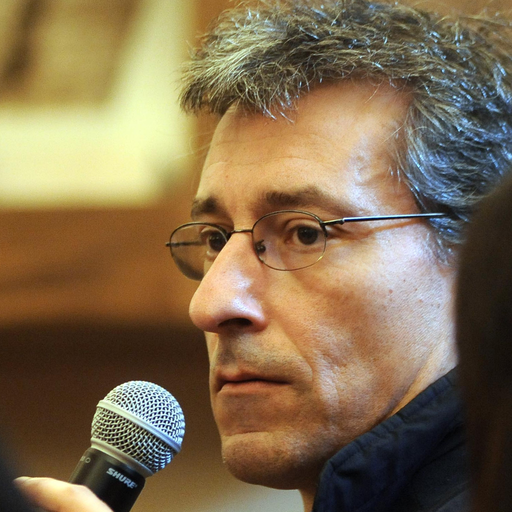}
    \end{subfigure} &
    \begin{subfigure}[b]{\qresultstextwidthzero\textwidth}
        \centering
        \includegraphics[width=\textwidth]{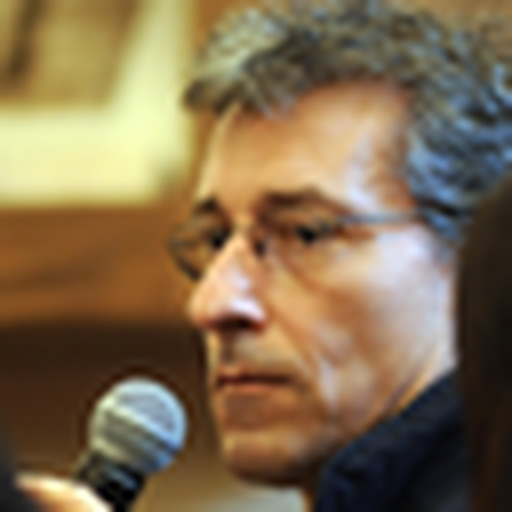}
    \end{subfigure} &
    \begin{subfigure}[b]{\qresultstextwidthzero\textwidth}
        \centering
        \includegraphics[width=\textwidth]{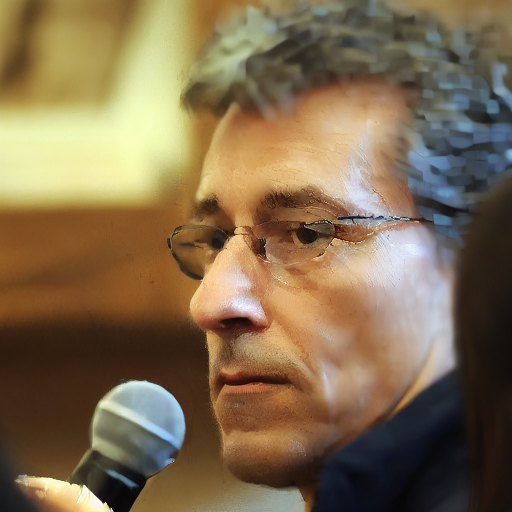}
    \end{subfigure} &
    \begin{subfigure}[b]{\qresultstextwidthzero\textwidth}
        \centering
        \includegraphics[width=\textwidth]{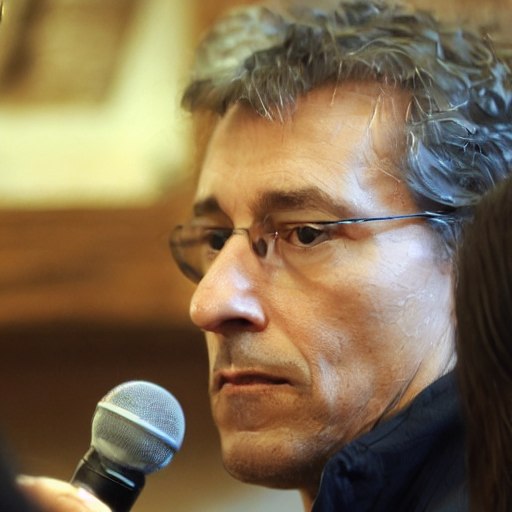}
    \end{subfigure} \\
    \begin{subfigure}[b]{\qresultstextwidthzero\textwidth}
        \centering
        \includegraphics[width=\textwidth]{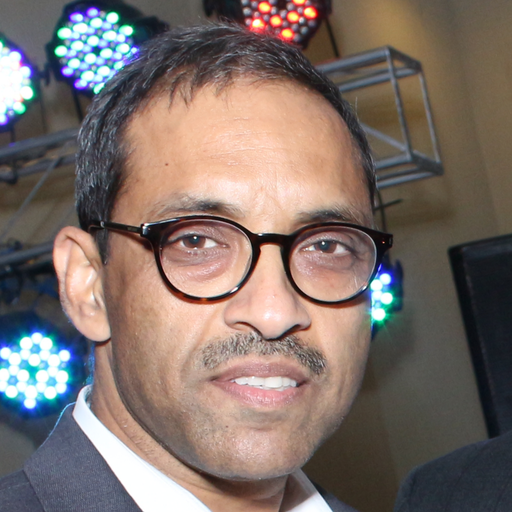}
    \end{subfigure} &
    \begin{subfigure}[b]{\qresultstextwidthzero\textwidth}
        \centering
        \includegraphics[width=\textwidth]{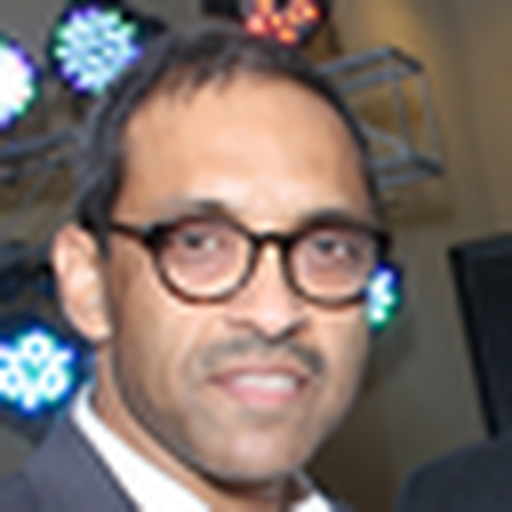}
    \end{subfigure} &
    \begin{subfigure}[b]{\qresultstextwidthzero\textwidth}
        \centering
        \includegraphics[width=\textwidth]{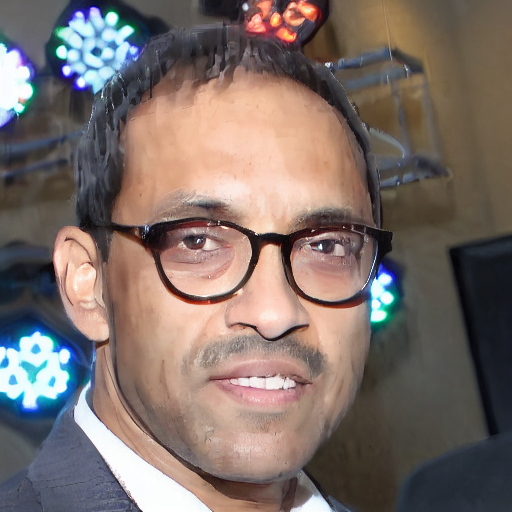}
    \end{subfigure} &
    \begin{subfigure}[b]{\qresultstextwidthzero\textwidth}
        \centering
        \includegraphics[width=\textwidth]{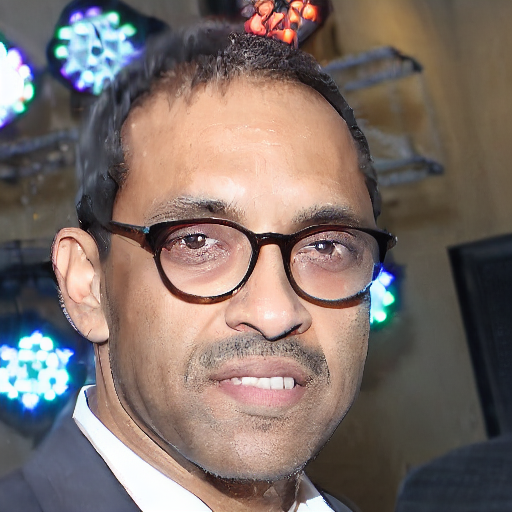}
    \end{subfigure} \\
    \begin{subfigure}[b]{\qresultstextwidthzero\textwidth}
        \centering
        \includegraphics[width=\textwidth]{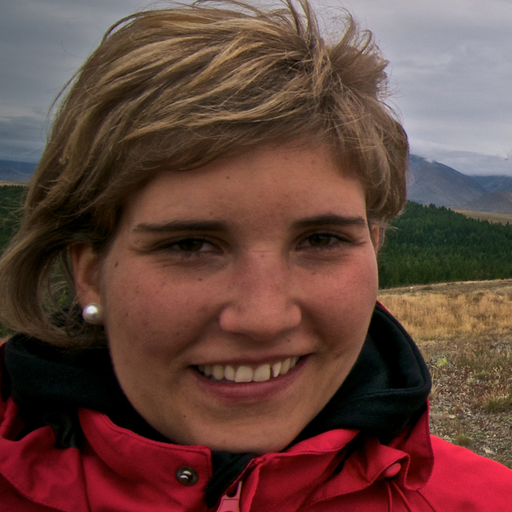}
    \end{subfigure} &
    \begin{subfigure}[b]{\qresultstextwidthzero\textwidth}
        \centering
        \includegraphics[width=\textwidth]{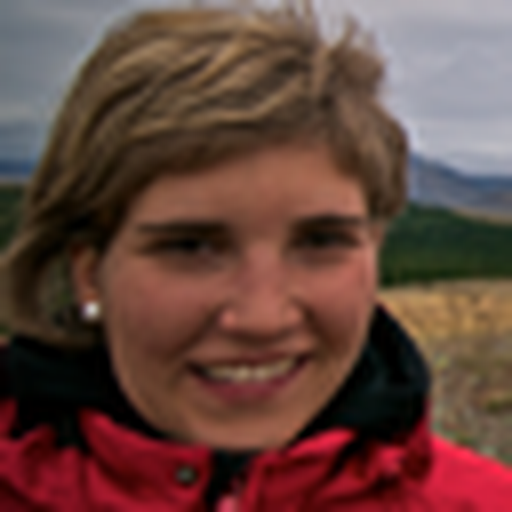}
    \end{subfigure} &
    \begin{subfigure}[b]{\qresultstextwidthzero\textwidth}
        \centering
        \includegraphics[width=\textwidth]{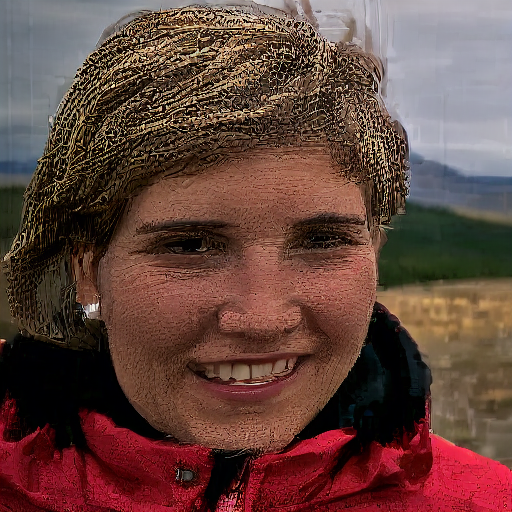}
    \end{subfigure} &
    \begin{subfigure}[b]{\qresultstextwidthzero\textwidth}
        \centering
        \includegraphics[width=\textwidth]{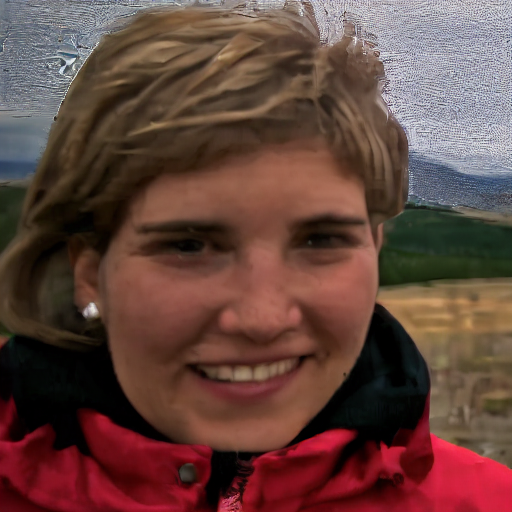}
    \end{subfigure}
    \end{tabular}
    \caption{Qualitative results for the super-resolution task on the FFHQ-1K ($512\times512$) dataset. For the first sample, PFLD-10 suffers from fewer artifacts than PSLD. For the second sample, PFLD-10 creates a slightly sharper output than PSLD. For the third sample, PSLD leaves a considerable amount of noise in the hair and face region and the PFLD-10 output is visually smoother.}
    \label{fig:qualitative_results_5}
\end{figure*}

\newcommand{\qresultsgaussianffhqa}{00306}
\newcommand{\qresultsgaussianffhqb}{00336}
\newcommand{\qresultsgaussianffhqc}{00372}

\begin{figure*}[htbp]
    \begin{tabular}{cccc}
        Ground Truth & Corrupted Image & PSLD & PFLD-10 (ours) \\
    \centering
    \begin{subfigure}[b]{\qresultstextwidthzero\textwidth}
        \centering
        \includegraphics[width=\textwidth]{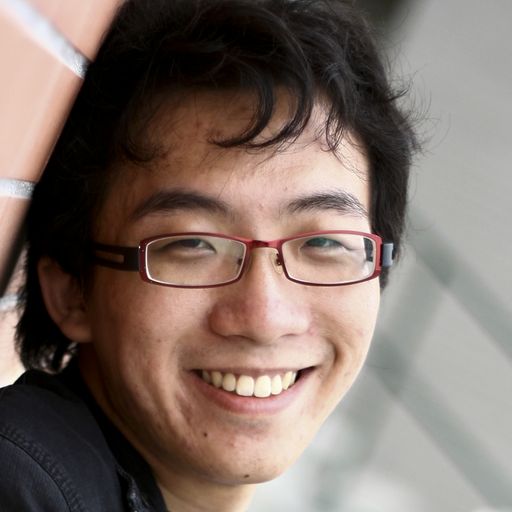}
    \end{subfigure} &
    \begin{subfigure}[b]{\qresultstextwidthzero\textwidth}
        \centering
        \includegraphics[width=\textwidth]{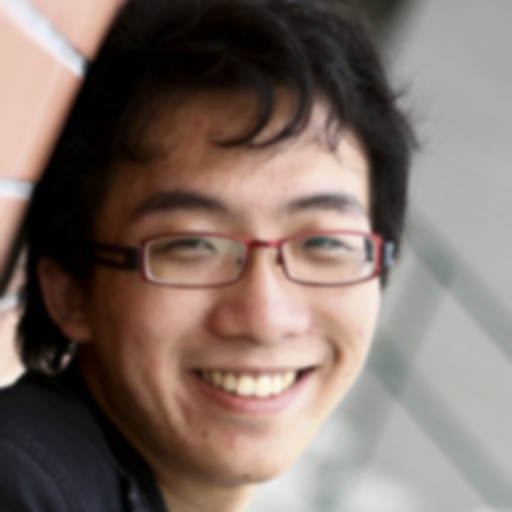}
    \end{subfigure} &
    \begin{subfigure}[b]{\qresultstextwidthzero\textwidth}
        \centering
        \includegraphics[width=\textwidth]{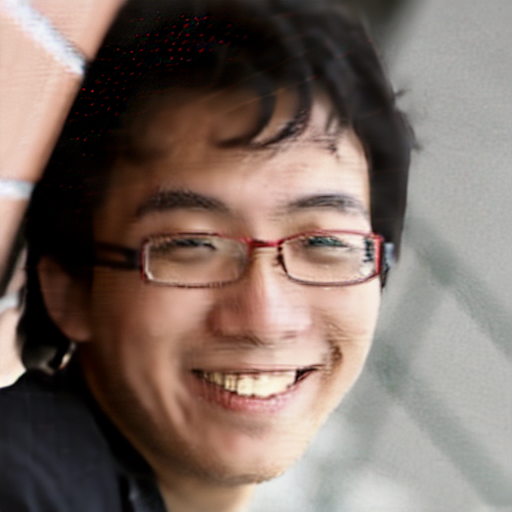}
    \end{subfigure} &
    \begin{subfigure}[b]{\qresultstextwidthzero\textwidth}
        \centering
        \includegraphics[width=\textwidth]{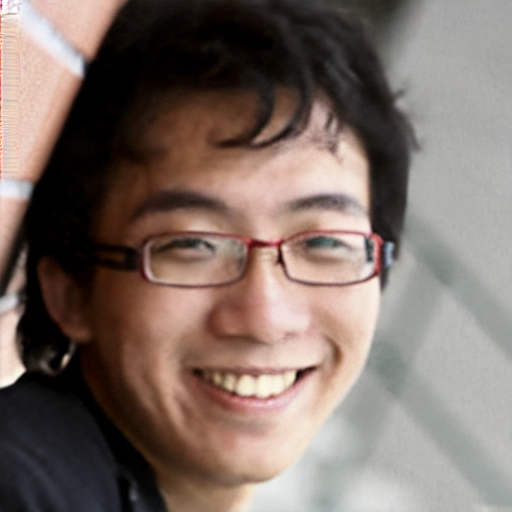}
    \end{subfigure} \\
    \begin{subfigure}[b]{\qresultstextwidthzero\textwidth}
        \centering
        \includegraphics[width=\textwidth]{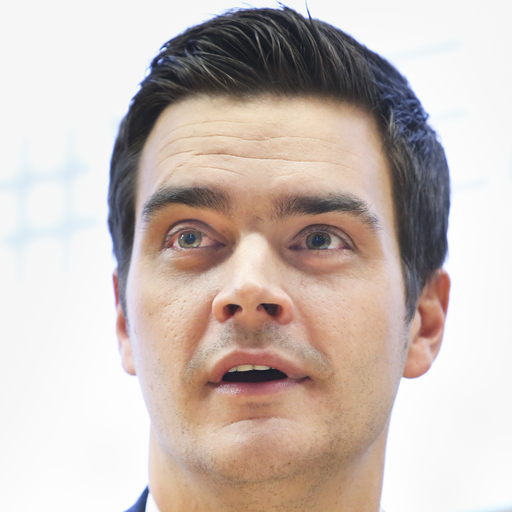}
    \end{subfigure} &
    \begin{subfigure}[b]{\qresultstextwidthzero\textwidth}
        \centering
        \includegraphics[width=\textwidth]{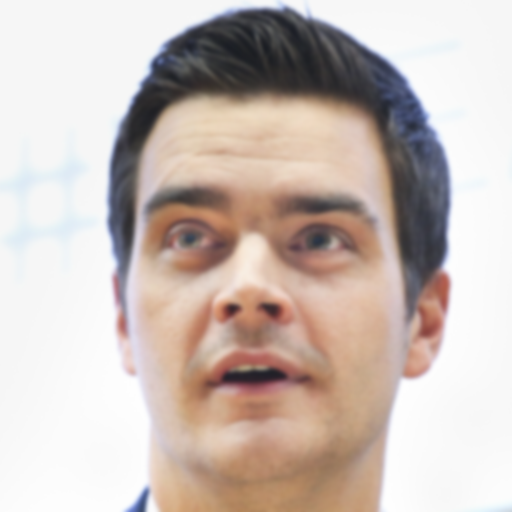}
    \end{subfigure} &
    \begin{subfigure}[b]{\qresultstextwidthzero\textwidth}
        \centering
        \includegraphics[width=\textwidth]{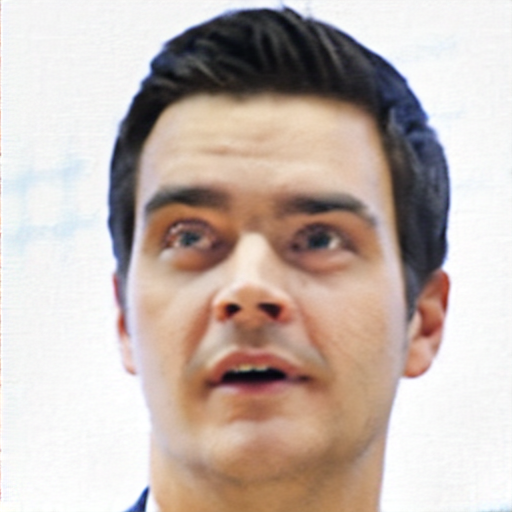}
    \end{subfigure} &
    \begin{subfigure}[b]{\qresultstextwidthzero\textwidth}
        \centering
        \includegraphics[width=\textwidth]{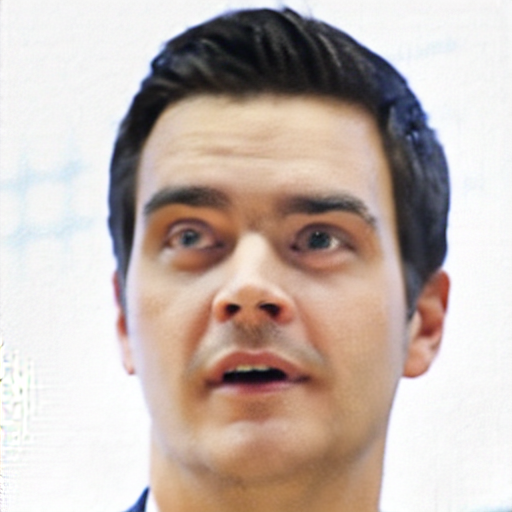}
    \end{subfigure} \\
    \begin{subfigure}[b]{\qresultstextwidthzero\textwidth}
        \centering
        \includegraphics[width=\textwidth]{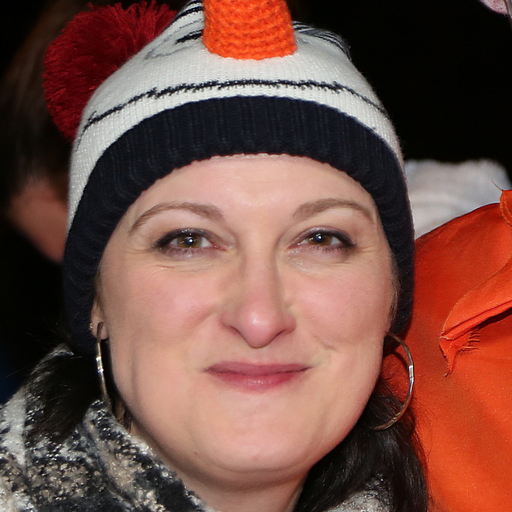}
    \end{subfigure} &
    \begin{subfigure}[b]{\qresultstextwidthzero\textwidth}
        \centering
        \includegraphics[width=\textwidth]{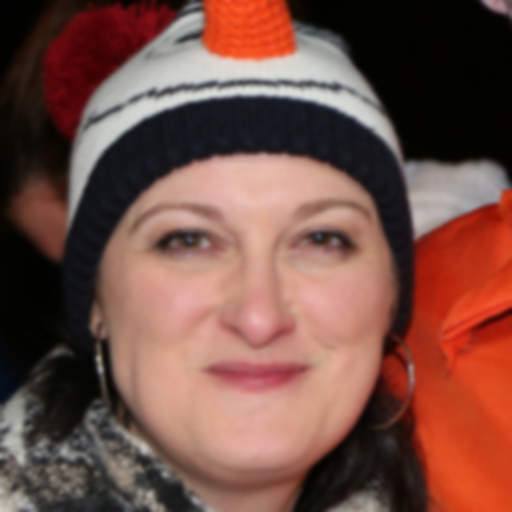}
    \end{subfigure} &
    \begin{subfigure}[b]{\qresultstextwidthzero\textwidth}
        \centering
        \includegraphics[width=\textwidth]{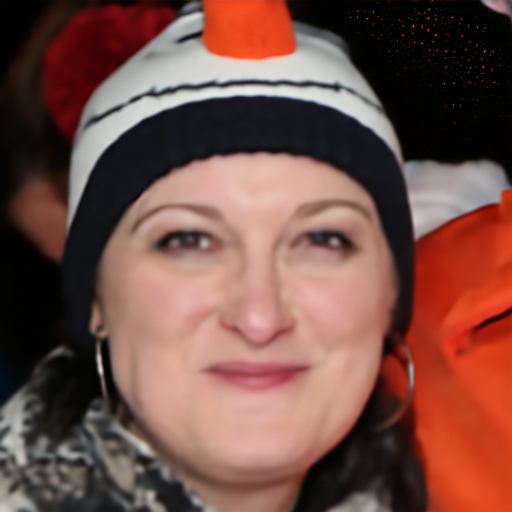}
    \end{subfigure} &
    \begin{subfigure}[b]{\qresultstextwidthzero\textwidth}
        \centering
        \includegraphics[width=\textwidth]{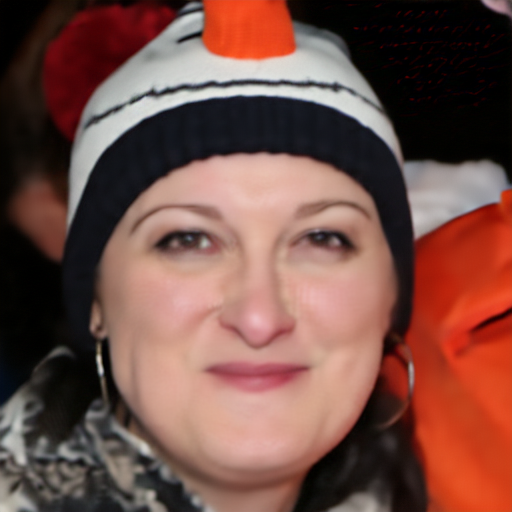}
    \end{subfigure}
    \end{tabular}
    \caption{Qualitative results for the Gaussian deblurring task on the FFHQ-1K ($512\times512$) dataset. For all samples presented, PFLD-10 produces outputs that are slightly closer to the ground truth. In the first sample, PFLD-10 did not produce an oversharpened output like PSLD did. In the second sample, the grainy, overly sharp artifacts around the right eyelid of the person are not present in the PFLD-10 output. In the final sample, PFLD-10 reconstructs fine detail around the left eyelid of the subject.}
    \label{fig:qualitative_results_6}
\end{figure*}


\EOD

\end{document}